\documentclass[10pt,twocolumn,letterpaper]{article}

\usepackage{iccv}
\usepackage{times}
\usepackage{epsfig}
\usepackage{graphicx}
\usepackage{amsmath}
\usepackage{amssymb}

\usepackage{siunitx}
\usepackage{cite}
\usepackage{ctable}
\usepackage{hhline}
\usepackage{booktabs}
\usepackage{subcaption}
\usepackage{csquotes}
\usepackage[pagebackref=true,breaklinks=true,letterpaper=true,colorlinks,bookmarks=false]{hyperref}

\iccvfinalcopy

\ificcvfinal\fi

\usepackage{float}

\newcommand{\DatasetNameGHI}{\emph{GHI}\xspace}
\newcommand{\DatasetNameLAION}{\emph{LAION-Human}\xspace}
\newcommand{\DatasetNameHumanArt}{\emph{Human-Art}\xspace}
\newcommand{\ModelName}{\emph{HumanSD}\xspace}
\newcommand{\LossName}{\emph{heatmap-guided denoising loss}\xspace}

\usepackage{url}

\usepackage{capt-of,etoolbox}

\makeatletter
\apptocmd\@maketitle{{\myfigure{}\par}}{}{}
\makeatother

\begin{document}

\title{HumanSD: A Native Skeleton-Guided Diffusion Model \\for Human Image Generation}

\author{
Xuan Ju$^{1,2}$\thanks{Equal contribution. $^{\ddag}$ Work done during an internship at IDEA.}~~$^{\ddag}$, 
Ailing Zeng$^{1*}$, 
Chenchen Zhao$^{2*}$, 
Jianan Wang$^{1}$, 
Lei Zhang$^{1}$, 
Qiang Xu$^{2}$\thanks{Corresponding author.}\\
$^{1}$International Digital Economy Academy, $^{2}$The Chinese University of Hong Kong \\
{\tt\small \{xju22, qxu\}@cse.cuhk.edu.hk, \{zengailing, wangjianan, leizhang\}@idea.edu.cn}\\
\url{https://idea-research.github.io/HumanSD/}
}

\ificcvfinal\thispagestyle{empty}\fi

\newcommand\myfigure{%
\vspace{-0.8cm}
\centering
   \includegraphics[width=0.89\linewidth,trim={4pt 4pt 4pt 4pt}]{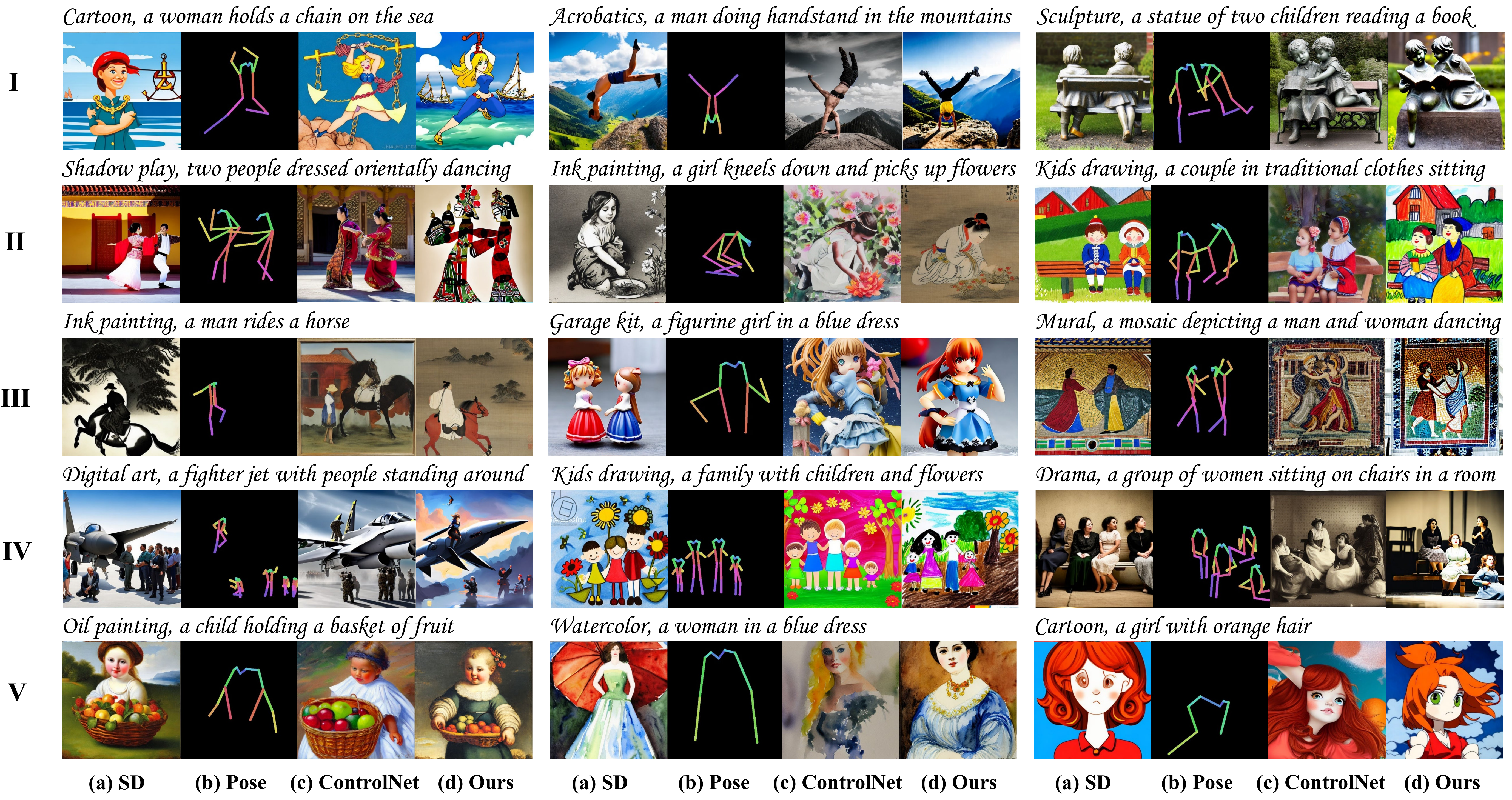}
\vspace{-0.3cm}
\captionof{figure}{This paper highlights multi-scenario human-centric image generation with precise pose control. Each group of displayed images includes: (a) a generation by the pre-trained pose-less text-guided stable diffusion (SD)~\cite{ldm22}, (b) pose skeleton images as the condition to ControlNet and our proposed \ModelName, (c) a generation by ControlNet~\cite{controlnet23}, and (d) a generation by \ModelName (ours). ControlNet and \ModelName receive both text and pose conditions. \ModelName shows its superiorities in terms of \textbf{(\uppercase\expandafter{\romannumeral1})} challenging poses, \textbf{(\uppercase\expandafter{\romannumeral2})} accurate painting styles, \textbf{(\uppercase\expandafter{\romannumeral3})} pose control capability, \textbf{(\uppercase\expandafter{\romannumeral4})} multi-person scenarios, and \textbf{(\uppercase\expandafter{\romannumeral5})} delicate details. \textbf{Best viewed with zoom-in}.}
\label{fig:teaser}
\vspace{1em}
}
\maketitle

\begin{abstract}
\vspace{-10pt}
Controllable human image generation (HIG) has numerous real-life applications. 
State-of-the-art solutions, such as ControlNet and T2I-Adapter, introduce an additional learnable branch on top of the frozen pre-trained stable diffusion (SD) model, which can enforce various conditions, including skeleton guidance of HIG.
While such a plug-and-play approach is appealing, the inevitable and uncertain conflicts between the original images produced from the frozen SD branch and the given condition incur significant challenges for the learnable branch, which essentially conducts image feature editing for condition enforcement.

In this work, we propose a \textbf{native} skeleton-guided diffusion model for controllable HIG called \textbf{\ModelName}. Instead of performing image editing with dual-branch diffusion, we fine-tune the original SD model using a novel \textbf{\LossName}. This strategy effectively and efficiently strengthens the given skeleton condition during model training while mitigating the catastrophic forgetting effects. \ModelName is fine-tuned on the assembly of three large-scale human-centric datasets with text-image-pose information, two of which are established in this work. As shown in Figure \ref{fig:teaser}, \ModelName outperforms ControlNet in terms of accurate pose control and image quality, particularly when the given skeleton guidance is sophisticated.

~

\vspace{-1cm}

\end{abstract}

\vspace{-0.2cm}
\section{Introduction}\label{sec:Introduction}

Controllable human image generation (HIG) aims to generate human-centric images under given conditions such as human pose~\cite{poseguided17_gan,nted22_gan,finegrained21_vae}, body parsing~\cite{zhu2017be,yu2019vtnfp}, and text~\cite{verbalpersonnets22, tips22, sharedspace21}. It has numerous applications (e.g., animation/game production~\cite{pan2022synthesizing} and virtual try-on~\cite{zhou2022cross}),  attracting significant attention from academia and industry.

While earlier controllable HIG solutions based on generative adversarial networks (GANs)~\cite{gan,poseguided17_gan,decomposed20_gan,mustgan21_gan,pise21_gan,spgnet21_gan,ctnet21_gan,nted22_gan,dptn22_gan} and variational auto-encoders (VAEs)~\cite{vae,transformation20_vae,finegrained21_vae,kpe22_vae} have been successfully applied in certain applications (e.g., virtual try-on), they have not gained mainstream acceptance due to their training difficulties and poor multi-modality fusion and alignment capabilities~\cite{zhang2022humandiffusion}. Recently, diffusion models~\cite{ddim, ddpm,ldm22} have demonstrated unprecedented text-to-image generation performance~\cite{ramesh2022hierarchical} and quickly become the dominant technique in this exciting field. However, it is difficult to provide precise position control with text information, especially for deformable objects such as humans.

To tackle the above problem, two concurrent controllable diffusion models were proposed in the literature: ControlNet~\cite{controlnet23} and T2I-Adapter~\cite{t2i23}. Both models introduce an additional learnable diffusion branch on top of the frozen pre-trained stable diffusion (SD) model~\cite{ldm22}. The additional branch enables the enforcement of various conditions such as skeleton and sketch during image generation, which greatly improves the original SD model in terms of controllability, thereby gaining huge traction from the community.

However, the learnable branch in such dual-branch diffusion models is essentially performing a challenging image feature editing task and suffers from several limitations. Consider the skeleton-guided controllable HIG problem that generates humans with specific poses. Given text prompts containing human activities, the SD branch may generate various images that are inconsistent with the skeleton guidance, e.g., humans could present at different places with various poses. Therefore, the extra condition branch needs to learn not only how to generate humans according to the given skeleton guidance but also how to suppress various inconsistencies, making training more challenging and inference less stable. Generally speaking, the larger the gap between skeleton guidance and original images produced by the frozen SD branch, the higher discrepancy between the given guidance and generated human images. Moreover, the inference cost of these dual-branch solutions largely increases compared to the original SD model.

\begin{figure*}[htbp]
    \centering
    \includegraphics[width=0.99\linewidth]{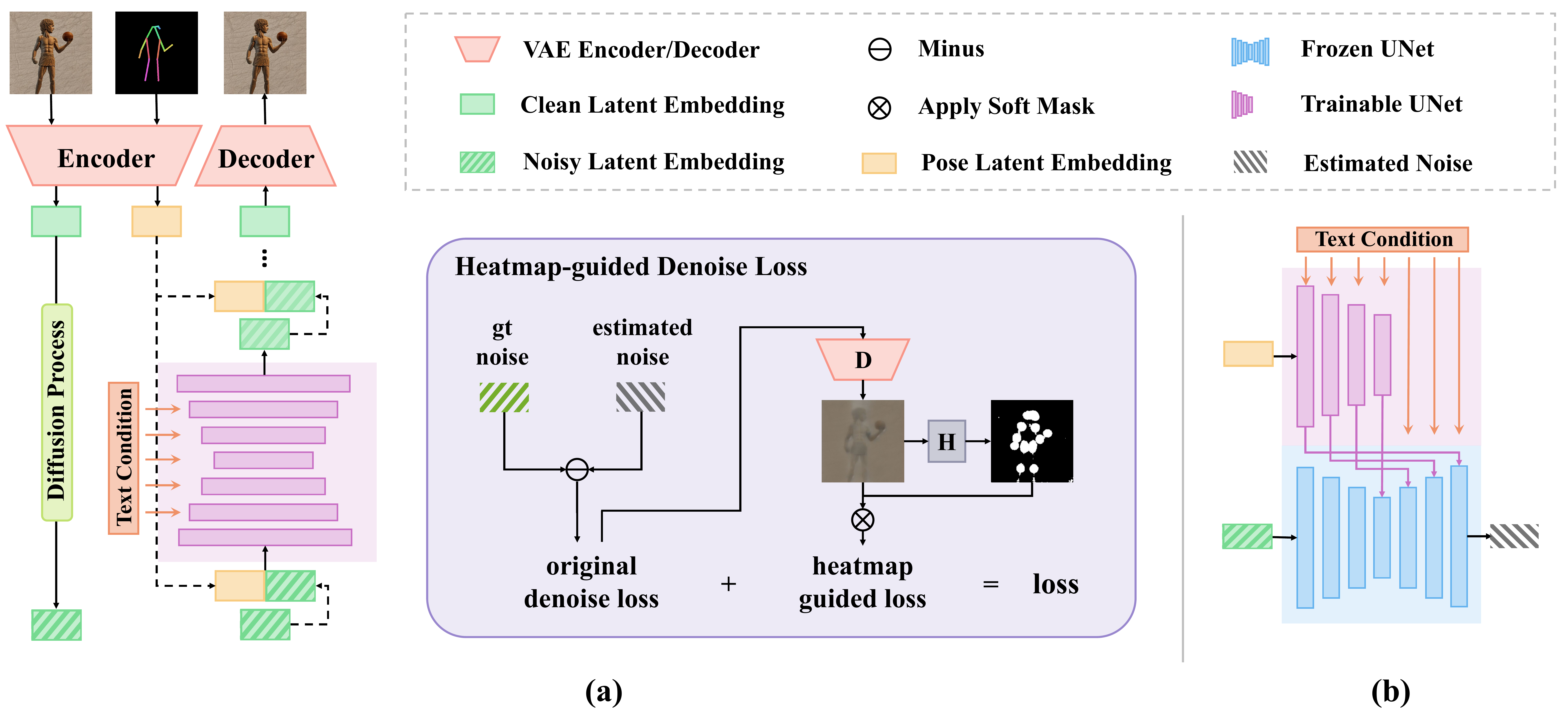}
        \vspace{-0.2cm}
    \caption{\textbf{Overview}. \textbf{(a)} shows \textbf{the proposed framework \ModelName} with a novel \LossName. Given the pose condition, our model inputs the corresponding skeletal image into the VAE encoder of a pre-trained SD to get the pose latent embedding. The embedding is then concatenated with the noisy latent embedding generated by diffusion and inputted into the UNet. In the training stage, the \LossName helps UNet concentrate on the specific areas humans are located in, especially when the human generation has poor performance, by increasing its weight in the loss function. \textbf{(b)} shows the recent SOTA method, \textbf{ControlNet}~\cite{controlnet23}, which doubles an SD UNet encoder for condition extraction and freezes the original SD branch to maintain image generation ability.
    } 
    \vspace{-0.4cm}
\label{fig:main_graph}
\end{figure*}

In contrast to employing an additional trainable branch for controllable HIG, this work proposes a \textbf{native} skeleton-guided diffusion model, named \ModelName. By directly fine-tuning the SD model~\cite{ldm22} with skeleton conditions concatenated to the noisy latent embeddings, as shown in Figure \ref{fig:main_graph} (a), \ModelName~can natively guide image generation with the desired pose, instead of conducting a challenging image editing task. To mitigate the catastrophic forgetting effects caused by model overfitting during fine-tuning, we propose a novel \LossName~for diffusion models to disentangle between conditioned humans and unconditioned backgrounds in the training stage. Such a disentanglement forces the fine-tuning process to concentrate on the generation of foreground humans while minimizing unexpected overrides of the pre-trained SD model parameters that hurt the model's generation and generalization abilities.

Besides the algorithm, training data is another important factor determining model performance~\cite{schuhmann2021laion}. To improve the HIG quality of \ModelName, we fine-tune our model on three large-scale human-centric datasets containing high-quality images and the corresponding 2D skeletal information and text descriptions: \DatasetNameGHI, \DatasetNameLAION, and \DatasetNameHumanArt. Specifically, \DatasetNameGHI and \DatasetNameLAION are established in this work. \DatasetNameGHI has 1M multi-scenario images generated from SD with crafted prompts, and only the top 30\% with the highest image quality are selected. For \DatasetNameLAION, it selects 1M human-centric images from the LAION-Aesthetics~\cite{schuhmann2022laion} via filtering.

The main contributions of this work include: 
\begin{itemize}
\vspace{-0.15cm}
    \item We propose a new HIG framework \ModelName with a novel \LossName, to natively generate human images with highly precise pose control yet no extra computational costs during inference.
\vspace{-0.15cm}
    \item We introduce two large-scale human-centric datasets with a standard development process, which facilitates multi-scenario HIG tasks with large quantities, rich data distribution, and high annotation quality.
\vspace{-0.15cm}
    \item To demonstrate the effectiveness and efficiency of \ModelName, we apply a series of evaluation metrics covering image quality, pose accuracy, text-image consistency, and inference speed to compare our model with previous works in a fair experimental setting.
\end{itemize}
\vspace{-0.15cm}

With the above, \ModelName outperforms state-of-the-art solutions such as ControlNet regarding pose control and human image generation quality, particularly when the given skeleton guidance is sophisticated.

\section{Related Work}

\subsection{Pose-Guided Human Image Generation}

During the past two decades, pose-guided controllable HIG~\cite{poseguided17_gan,decomposed20_gan,mustgan21_gan,pise21_gan,spgnet21_gan,ctnet21_gan,nted22_gan,dptn22_gan,transformation20_vae,finegrained21_vae,kpe22_vae} has gained lots of attention in academia and industry due to the pose's validity in motion description~\cite{coco,zeng2022smoothnet,zeng2022deciwatch,zeng2021learning,lin2023osx,yang2023explicit}.
With source images and pose conditions (e.g., skeletal images or body parsing), pose-guided HIG models output photorealistic images with source images' appearance and desired poses.
These algorithms are mainly based on GANs~\cite{gan,poseguided17_gan,decomposed20_gan,mustgan21_gan,pise21_gan,spgnet21_gan,ctnet21_gan,nted22_gan,dptn22_gan} or VAEs~\cite{vae,transformation20_vae,finegrained21_vae,kpe22_vae}.
Exclusively focusing on natural-scene manipulation, they fail in diverse cross-modality feature alignment due to limitations in model design, inappropriate condition injection strategies, and lack of diversity in training data, which lead to unrealistic and poor results with artificial scenario source images or arbitrary pose inputs. 
In addition, these models strongly depend on the source-target paired images that are hard to acquire and lack diversity.

Different from images, text has become a flexible, user-friendly, and informative condition with the rise of large vision-language models~\cite{clip}.
Some works involve text conditions to guide HIG but are limited to small-scale vocabulary pools and fail with open vocabulary~\cite{verbalpersonnets22,sharedspace21,tips22}.
Among the very recent works, ControlNet~\cite{controlnet23}, T2I-Adapter~\cite{t2i23}, and GLIGEN~\cite{gligen23} introduce methods of adding arbitrary conditions. ControlNet and T2I-Adapter add additional trainable modules to pre-trained text-to-image diffusion models~\cite{ldm22}. 
The target of designing general frameworks makes them not well-targeted to humans that appear with diverse poses, fine-grained body parts, styles, viewpoints, sizes, and quantities. Moreover, their models suffer from trainable-frozen branch conflicts, thus showing inadequate pose control ability. Superior to previous work, \ModelName~is efficient as well as high-precision in human pose control, and specially designed for open-world multi-scenario HIG.

\subsection{Human Image Generation Datasets}

Current HIG datasets such as iDesigner~\cite{idesigner}, 
DeepFashion~\cite{deepfashion}, 
Market1501~\cite{market1501}, and MSCOCO~\cite{coco} mainly focus on the real-scene human generation and provide noisy paired source-target images.  
These mainstream datasets have limited scenarios (e.g., dress-up, street photography), and are not generalizable to other scenarios such as cartoons, oil paintings, and sculptures.

Recently, \DatasetNameHumanArt~\cite{humanart} provides 50K human-centric images in five natural and fifteen artificial scenes with precise pose and text annotations. Specially designed for multi-scenario human-centric tasks, \DatasetNameHumanArt is suitable for validating the quality and diversity of existing generation methods. However, the limited data scale of \DatasetNameHumanArt makes it inadequate for large model training. Laion-5B~\cite{schuhmann2022laion} is a publicly available dataset with sufficient text-image paired data but contains many human irrelevant images. 
ControlNet~\cite{controlnet23} adopts the human pose estimator OpenPose~\cite{cao2017openpose} on internet-scratched images to collect 200K pose-image-text pairs, most of which are real-scene images. Using these data pairs in training will lead to a significant distribution bias towards real yet low-diversity scenes.

This work provides a standard development process for large-scale multi-scenario text-image-pose datasets targeted at skeleton-guided HIG, which addresses the absence of suitable training and testing datasets.

\section{Preliminaries and Motivation}

This section introduces the details of the conflicts in recent SOTA SD-based HIG methods before outlining the motivation for designing \ModelName in section~\ref{sec:pose_image_condition_addition}.
These methods, notably ControlNet and T2I-Adapter, use the Latent Diffusion Model (LDM~\cite{ldm22}) as the foundation for its high trainability and high-generation quality, which will be introduced in Section \ref{sec:Preliminary_for_Latent_Diffusion_Model}. Then, Section \ref{sec:double_sd_conflict} states the conflicts in ControlNet and T2I-Adapter. Since these two models have a similar design, we take ControlNet as an example.

\subsection{Preliminaries - The Latent Diffusion Model}
\label{sec:Preliminary_for_Latent_Diffusion_Model}
LDM, more known as Stable Diffusion (SD), is a diffusion model\cite{diffusion2015} 
conducted on latent embeddings instead of images.
Images are projected into latent embeddings by a VAE and then guided by text conditions in the latent space.
LDM has a latent-space loss function with a similar form to vanilla diffusion models\cite{ddpm,ddim}:

\vspace{-0.55cm}
\begin{equation}
\small
L_\text{LDM}=\mathop \mathbb{E} \limits_{t,z,\epsilon}\left[ \left\| \epsilon -\epsilon _{\theta}\left( \sqrt{\bar{\alpha}_t}z_0+\sqrt{1-\bar{\alpha}_t}\epsilon, c, t \right) \right\| ^2 \right]
\label{eq:ldm_loss}
\vspace{-0.2cm}
\end{equation}
where $z_0$ is the latent embedding of a training sample $\rm{x}_0$; $\epsilon_\theta$ and $\epsilon$ are respectively the noise estimated by the UNet $\theta$ and the ground truth noise injected at the corresponding diffusion timestep $t$; $c$ is the embedding of all conditions involved in the generation; $\bar{\alpha}_t$ is the same coefficient as that in vanilla diffusion models.

\subsection{Conflicts in Dual-Branch Solution}

\label{sec:double_sd_conflict}

In this section, we provide more detailed theoretical analyses on ControlNet~\cite{controlnet23} condition addition strategy. We argue that the conflict between the behavior of the frozen image-generation branch and the trainable condition-injection branch results in the degradation of pose control.

\begin{figure}[htbp]
    \centering
    \includegraphics[width=0.95\linewidth]{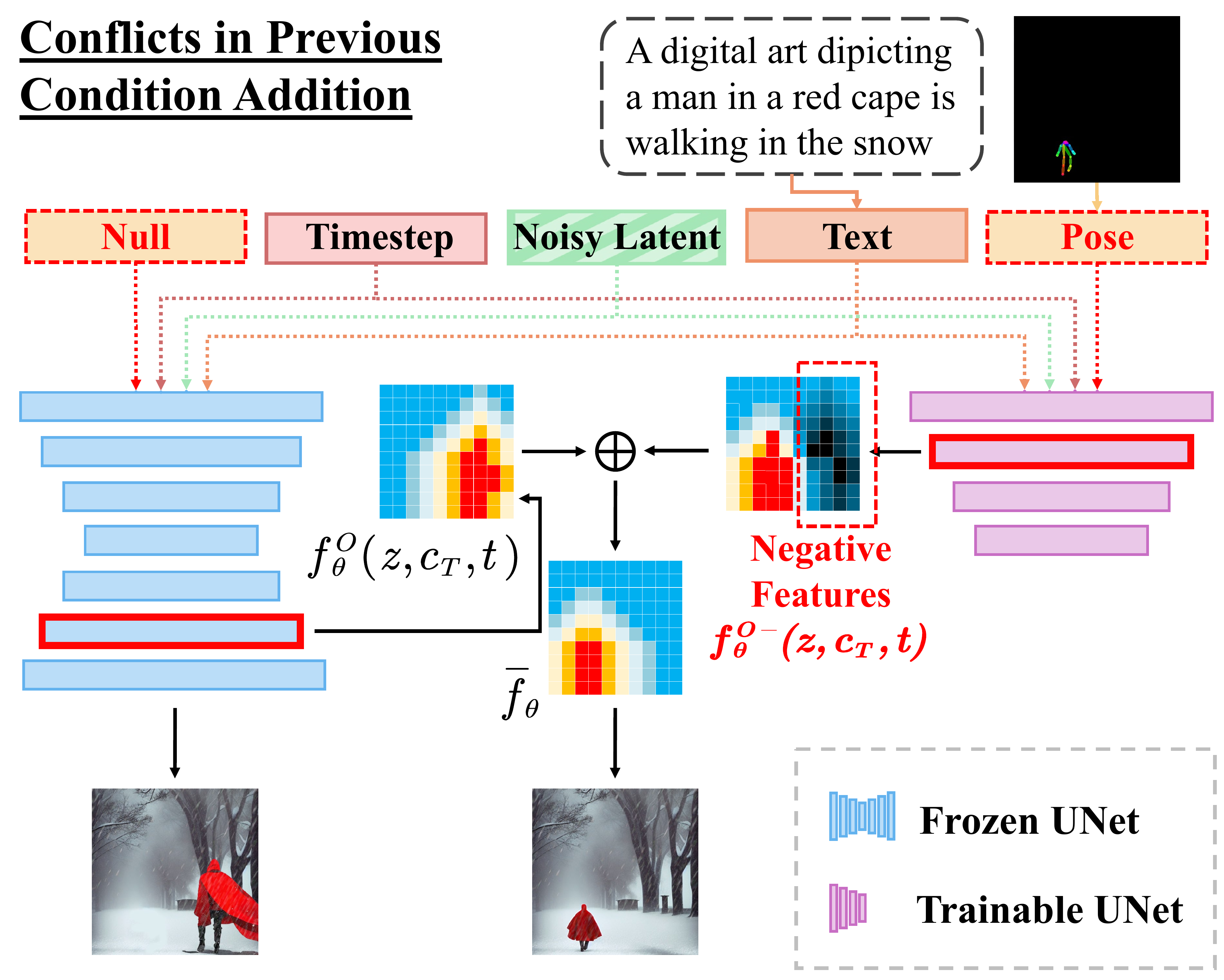}
    \caption{An example showing the \textbf{conflicts} in the behavior of the two branches in ControlNet~\cite{controlnet23}. }
    \label{fig:conflict}
    \vspace{-0.7cm}
\end{figure}

As shown in Figure~\ref{fig:conflict}, ControlNet is a plug-and-play approach for conditional image generation. It clones an SD branch to extract hierarchical features from the added condition and freezes the original SD branch to preserve generation ability. The trainable and frozen neural network
blocks are connected with a convolution layer. The convolution layer takes trainable features as input and its output is added to the frozen features. 

We denote the feature in the UNet of the original and the additional SD branches as $f^O_{\theta}(z, c_T, t)$ and $f^A_{\theta}(z, c, t)$, where $c_T$ is the text condition, $c=c_T+c_P$ is the ensemble of $c_T$ and pose condition $c_P$. Note that noise $\epsilon^O_{\theta}(z, c_T, t)$ and $\epsilon^A_{\theta}(z, c, t)$ can be viewed as the feature output by the last UNet layer. As shown in Figure \ref{fig:conflict}, $f^O_{\theta}(z, c_T, t)$ can be divided into a positive part $f^{O+}_{\theta}(z, c_T, t)$ and a negative part $f^{O-}_{\theta}(z, c_T, t)$, based on their consistency with $c_P$.

\vspace{-0.55cm}
\begin{equation}
f^O_{\theta}(z, c_T, t)=f^{O+}_{\theta}(z, c_T, t)+f^{O-}_{\theta}(z, c_T, t)
\label{eq:positive_and_negative_noise}
\vspace{-0.2cm}
\end{equation}

For dual-branch models with text-and-pose-guided generation,
an ideal estimated feature $\bar{f}_\theta$ should satisfy:

\vspace{-0.2cm}

\begin{equation}
\bar{f}_{\theta}=f_{\theta}^{O+}(z, c_T, t)+\tilde{f}_{\theta}^+(z, c, t),
\label{eq:ideal_noise}
\vspace{-0.2cm}
\end{equation}
where $\tilde{f}_{\theta}^+(z, c, t)$ ensures fine-grained pose control that cannot be guaranteed by $f_\theta^{O+}(z, c_T, t)$. We also have,

\vspace{-0.35cm}

\begin{equation}
\bar{f}_{\theta}=f_{\theta}^{O}(z,c_T,t)+f_{\theta}^{A}(z,c,t)
\label{eq:ideal_noise2}
\vspace{-0.2cm}
\end{equation}

Thus, we can obtain the feature in the additional  SD branches as follows:

\vspace{-0.35cm}

\begin{equation}
f^A_{\theta}(z, c, t)=\tilde{f}_{\theta}^+(z, c, t)-f^{O-}_{\theta}(z, c_T, t)
\label{eq:noise_surpress}
\vspace{-0.2cm}
\end{equation}

This leads to indirect noise generation during inference, where the additional (trainable) branch has to learn how to (1) identify the positive and negative parts of the estimated noise given the pose condition, (2) suppress the negative part, and (3) generate the extra positive part. The frozen SD branch results in a permanent existence of conflicts between the negative part and the extra positive part.
In contrast, for fine-tuning-based methods with all parameters trainable, the models go through a smooth and stable training process, and naturally learn to process the pose conditions and the cross-condition balance, thus avoiding the conflict.

\section{Method}
\label{sec:pose_image_condition_addition}

To resolve the conflict in previous SD-based methods, we introduce \ModelName, a native skeleton-guided diffusion model for precise and efficient multi-scenario human image generation. Vanilla fine-tuning faces the problem of catastrophic amnesia and over-fitting. To address this issue, 
we propose a condition addition strategy with a novel loss, which is illustrated in Section \ref{sec:model_condition_addition} and Section \ref{sec:new_loss}.  
Lastly, we provide a dataset construction process for multi-scenario Human-centric Image Generation in Section \ref{sec:dataset_process}.

\subsection{Skeleton Condition Addition}
\label{sec:model_condition_addition}

As shown in figure \ref{fig:main_graph} (a), our proposed \ModelName adds pose condition using a skeleton image with the same size as the input image, which provides explicit position information. 
In order to align the pose conditions with the latent embeddings of the input images, the skeleton image is then processed with the VAE encoder.
Different from text conditions, we do not add pose latent embedding with attention in each UNet block, but directly concatenate it to the noisy latent embeddings. This ensures that information on the same density level is processed at the same stage, which results in improved structure information integration.

\subsection{The Heatmap-guided Denoise Loss}
\label{sec:new_loss}

Fine-tuning deep neural networks with no protection can easily lead to catastrophic forgetting, where the performance of previous tasks drastically degrades when learning to perform a new task.
Directly fine-tuning diffusion models with new data and new conditions leads to the same problem (e.g., Anything Model~\cite{anything}, which is fine-tuned from SD to generate anime images, is unable to produce images in other styles).
Such performance degradation partially results from the non-discriminatory learning on all pixels of the image. This, to some extent, is reasonable for the conditions with global information (e.g., general text descriptions). However, for conditions with local structure information (e.g., pose condition with specific position information), fine-tuning the whole image results in a quality decline of condition-invariant regions (e.g., background).

\begin{figure}[htbp]
\vspace{-0.175cm}
    \centering
    \includegraphics[width=0.9\linewidth]{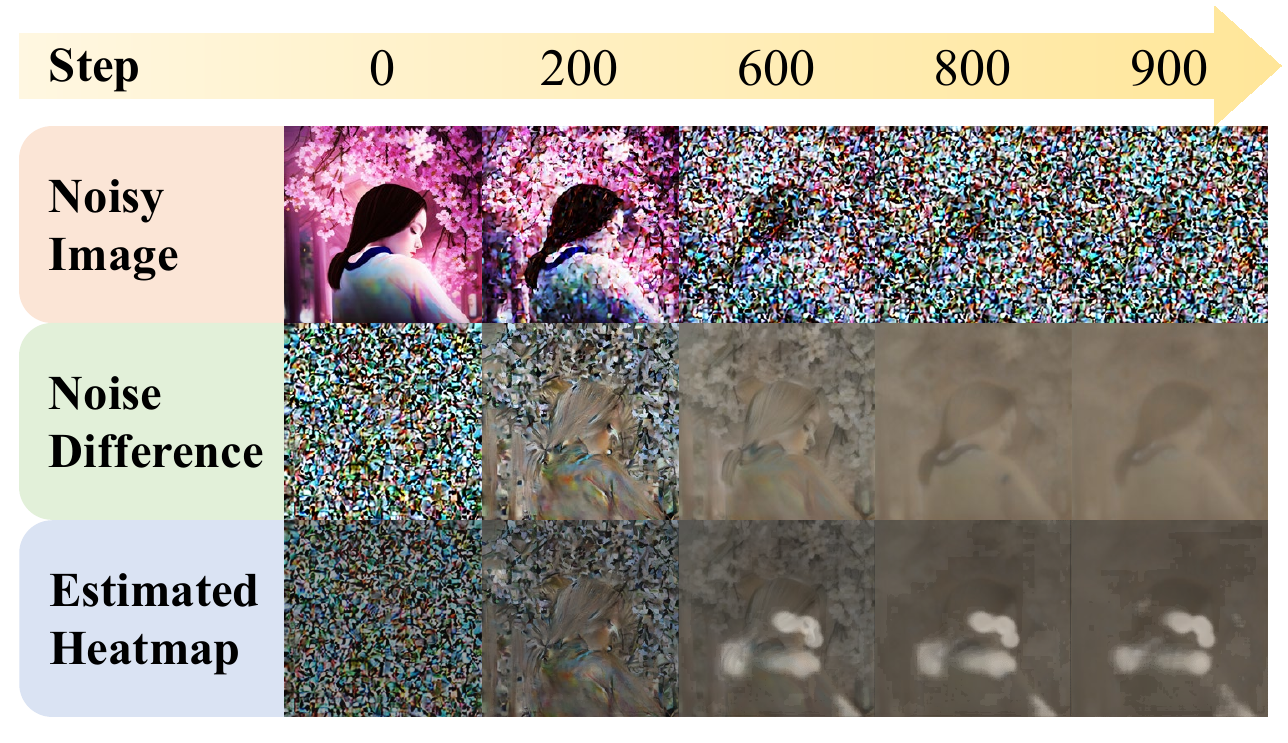}
    \vspace{-0.2cm}
    \caption{An illustration explaining \textbf{the calculation of $W_a$} in different diffusion steps.} 
\label{fig:noise_map}
\vspace{-0.225cm}
\end{figure}

To address this problem, we propose a \LossName~to fine-tune the diffusion model in a protection mode when adding a new structure-aware condition, which pays special attention to the training of the newly added condition and leaves the condition-invariant parts of the image to the pre-trained backbone, thus reaching high performance in both generation quality and condition-image consistency. 
The \LossName~takes effect by explicitly adding an aggregated heatmap weight $W_a$ to the original loss of the diffusion model. The loss function is then modified from Equation \ref{eq:ldm_loss} to Equation \ref{eq:attentional_loss}.

\vspace{-0.5cm}
\begin{equation}
\small
L_\text{h}= \mathop \mathbb{E} \limits_{t, z, \epsilon}\left[   \left\| W_a \cdot  \left( \epsilon -\epsilon _{\theta}\left( \sqrt{\bar{\alpha}_t}z_0+\sqrt{1-\bar{\alpha}_t}\epsilon, c, t \right)\right) \right\| ^2 \right] 
\label{eq:attentional_loss}
\vspace{-0.2cm}
\end{equation}

One of the most straightforward designs for $W_a$ is to assign bigger priority factors for feature pixels that are more related to the condition. However, diffusion is a step-by-step noise addition process, and 
not all steps are essential to condition injection. Therefore, assigning a constant weight map in all steps may disrupt the training process.

As a result, we need to (1) find out what the model respectively learns at different steps and stages, and (2) determine a weight function $W_a(t)$ based on the step-wise model behavior. The first row of Figure \ref{fig:noise_map} shows the decoded noisy latents in different steps; the second row shows the corresponding differences between the estimated noise and its ground truth (determined in the diffusion process), and the third row shows the corresponding heatmaps generated by a pre-trained human pose heatmap estimator~\cite{cheng2020higherhrnet} with the noise difference as inputs.
Using the heatmap as the description of $W_a$, the diffusion model can learn better with greater concentration on condition (human pose) processing. More detailed implementation of \LossName can be found in Figure \ref{fig:main_graph} (a).

\subsection{The Dataset Construction Process}
\label{sec:dataset_process}

\begin{figure}[htbp]
    \centering
    \includegraphics[width=1\linewidth]{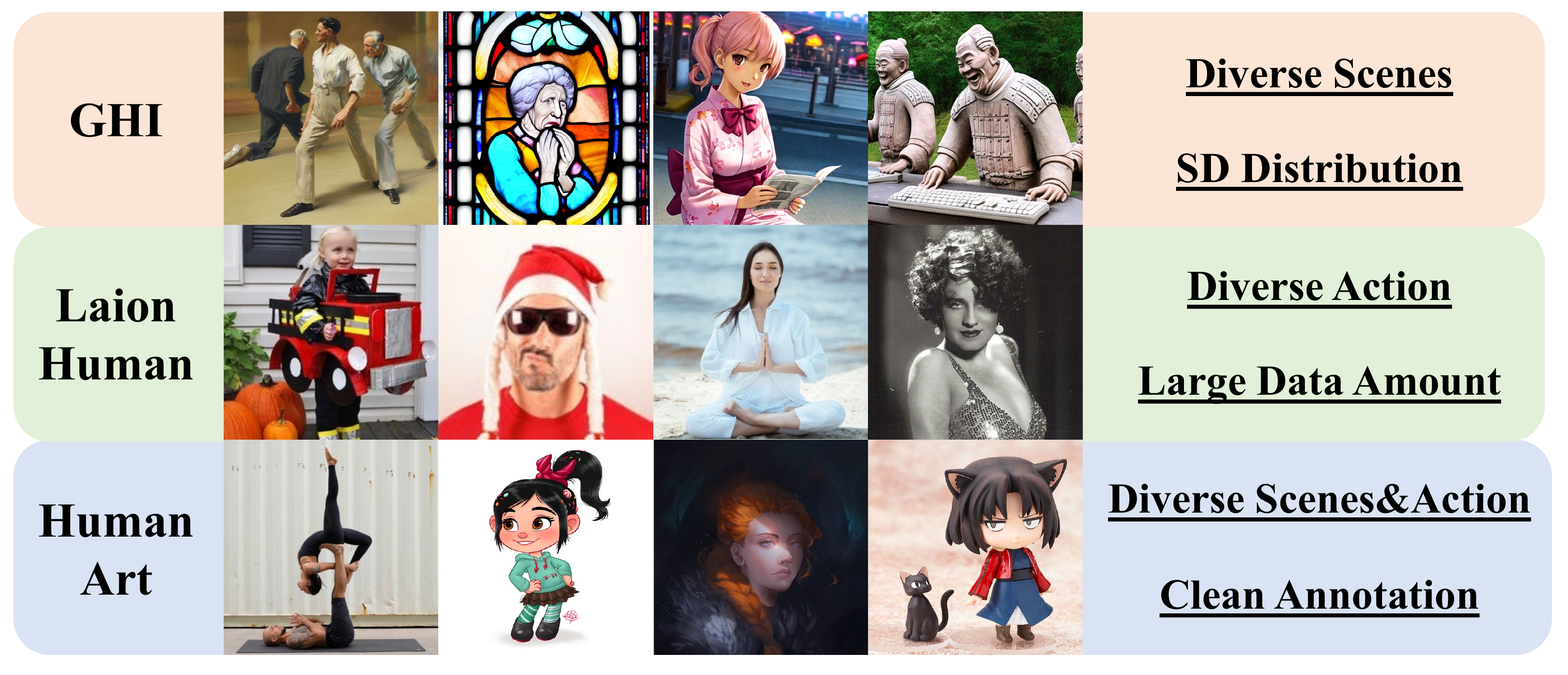}
    \caption{\textbf{Examples and characteristics} of the used datasets \DatasetNameGHI \& \DatasetNameLAION, and \DatasetNameHumanArt~\cite{humanart}.
    } 
    
\label{fig:dataset}
\end{figure}

Diffusion models require enormous amounts of data for training and fine-tuning. To ensure diverse data distribution in image scenes, human actions, and appearances, we introduce a standard dataset development process and construct $2$ large-scale datasets \DatasetNameGHI and \DatasetNameLAION. Figure \ref{fig:dataset} illustrates examples and characteristics of each dataset.

\textbf{\DatasetNameGHI}: It is an abbreviation for \textbf{G}enerated \textbf{H}uman \textbf{I}mages. Directly sampling data from SD's own learned distribution is a good way to maintain SD's generation capability with no new data distribution introduced. In order to maximize the exploitation of potential image possibilities in SD, we take advantage of prompt engineering~\cite{oppenlaender2022prompt,liu2022design} to design prompts that are constructed with 18 sub-prompt parts including image scene style, human number, human characteristics, action, and background descriptions (e.g., a realistic pixel art of two beautiful young girls running in the street of pairs at midnight, in winter, 64K, a masterpiece.). We use the pose estimator~\cite{cheng2020higherhrnet} trained on \DatasetNameHumanArt to detect character poses in diverse scenes. Then filter out images with the wrong human number, multi-arms and legs~\cite{kpe22_vae}, and low body integrity based on the detection results. The selection strategy ensures \DatasetNameGHI contains relatively clean annotations for text and pose, and increases image quality. This leads to a total number of 1M pose-image-text pairs that include 14 scenes (taken from \DatasetNameHumanArt) and 6826 human actions (taken from BABEL~\cite{BABEL}, NTU RGB+D 120~\cite{liu2019ntu}, HuMMan~\cite{cai2022humman}, HAA500~\cite{chung2021haa500}, and HAKE-HICO~\cite{li2019hake}) with one to three humans (with proportion of 7:2:1) in each image. 

\textbf{\DatasetNameLAION}: Similar with ControlNet~\cite{controlnet23} and T2I-Adapter~\cite{t2i23}, we construct a dataset \DatasetNameLAION containing large-scale internet images. Specifically, we collect about 1M image-text pairs from LAION-5B~\cite{schuhmann2022laion} filtered by the rules of high image quality and high human estimation confidence scores. Superior to ControlNet and T2I-Adapter, we adopt a versatile pose estimator trained on \DatasetNameHumanArt, which allows for selecting more diverse images such as oil paintings and cartoons. Importantly, \DatasetNameLAION contains more diverse human actions and more photorealistic images than data used in ControlNet and T2I-Adapter.

\textbf{\DatasetNameHumanArt}: \DatasetNameHumanArt~\cite{humanart} contains 50k images in 20 natural and artificial scenarios with clean annotation of pose and text, which can provide precise poses and multi-scenario for training and quantitative evaluation.
We follow \DatasetNameHumanArt's setting to divide the training and testing sets. 

Unless otherwise stated, we train \ModelName on the ensemble of \DatasetNameGHI, \DatasetNameLAION, and the training set of \DatasetNameHumanArt (denote as \emph{Union} in following sections), and test on the validation set of \DatasetNameHumanArt.

\section{Experiments}

\begin{table*}[htbp]
\small
\centering
\setlength{\tabcolsep}{0.4mm}{
\begin{tabular}{l|cc|cccc|c|c}
 \toprule
\multicolumn{1}{r|}{\textbf{Metrics}} & \multicolumn{2}{c|}{\textbf{Image Quality}} & \multicolumn{4}{c|}{\textbf{Pose Accuracy}}    & \textbf{Text-image Consistency}&\textbf{Inference Time} \\ 
  \textbf{Models}  & \textbf{FID} $\downarrow$ & \textbf{KID} $_{(\times 10)}$ $\downarrow$ &  \textbf{AP} $\uparrow$ & \textbf{AP(m)} $\uparrow$ & \textbf{CAP} $\uparrow$ & \textbf{PCE} $\downarrow$ & \textbf{CLIPSIM} $\uparrow$ & \textbf{Second per Image} $\downarrow$  \\ \midrule
Stable Diffusion & 41.55 & 2.99 &  0.09 & 0.00 & 49.05 & 1.76 & \textbf{33.86} & \textbf{3.88}\\

T2I-Adapter~\cite{t2i23} & 29.07 & 2.67 &  18.20 & 11.93 & 55.98 & 2.73 & 33.29 & 5.28 \\

ControlNet~\cite{controlnet23} & 27.24 & 2.60 & 18.45 & 11.71 & 57.18 & 2.47 & 33.16 & 6.37 \\

\ModelName(0.2M) & \textbf{26.28} & \textbf{2.56}  & \textbf{31.85} \textcolor{red}{$_{72.6\%\uparrow}$} &\textbf{24.95}\textcolor{red}{$_{109.1\%\uparrow}$} & \textbf{59.11}\textcolor{red}{$_{3.3\%\uparrow}$} & \textbf{1.61}\textcolor{red}{$_{34.8\%\downarrow}$}  &  32.98 & 3.89\\
\bottomrule
\end{tabular}}
\caption{\textbf{Quantitative comparisons between \ModelName and other SD-based models} (fair comparison). \ModelName is trained for around 300 GPU hours (95K iterations) on 0.2M text-image-pose pairs randomly selected from \DatasetNameLAION, similar to T2I-Adapter and ControlNet. Results demonstrate \ModelName's effectiveness and efficiency.}
\label{tab:all_metrics}
\vspace{-0.5cm}
\end{table*}

In this section, we validate that \ModelName outperforms currently SOTA SD-based (Section \ref{sec:compare_sd}) and GAN-based (Section \ref{sec:compare_gan}) methods on skeleton-guided HIG with 8 evaluation metrics explained in Section \ref{sec:evaluation_metrics}. Section \ref{sec:ablation} provides ablation studies on the \LossName, training datasets, and training iterations. Please refer to the supplementary material for implementation details.

\subsection{Evaluation Metrics}
\label{sec:evaluation_metrics}

To illustrate the effectiveness and efficiency of our proposed \ModelName, we use eight metrics covering four aspects: image quality, pose accuracy, text-image consistency, and inference time.

\textbf{Image Quality:} We report Fr\'echet Inception Distance (FID~\cite{fid}) and Kernel Inception Distance (KID~\cite{kid}), which are widely used to measure the quality of the syntheses. Specifically, we evaluate FID and KID on each \DatasetNameHumanArt scenario and report the mean value, which reflects both quality and diversity of the generation.

\textbf{Pose Accuracy:} We adopt distance-based Average Precision (AP)~\cite{coco}, Pose Cosine Similarity-based~\cite{posenet_similarity} AP (CAP) and People Count Error (PCE)~\cite{kpe22_vae}. These metrics measure the difference between the given pose condition and the pose result extracted from the generated image. Distance-based AP evaluates the keypoint-wise distances between the ground truth and the generated pose. We also provide AP(m) for medium-sized humans (with resolutions ranging from $32^2$ to $96^2$ following MSCOCO~\cite{coco}). We calculate CAP by simply replacing the distance error with the normalized cosine similarity error~\cite{posenet_similarity} in AP to evaluate the position-aligned similarity between the given pose and the generated pose. CAP eliminates the effect of absolute position and concentrates on pure action similarity. PCE measures the difference between the number of given skeletons and the generated humans. It effectively evaluates multi-person image generation, and partially reflects inconsistency~\cite{kpe22_vae} in single-person image generation, such as false numbers of heads, arms, and legs.

\vspace{5pt}
\textbf{Text-image Consistency:} The CLIP~\cite{clip} Similarity (CLIPSIM~\cite{clipsim}) evaluates text-image consistency between the generated images and corresponding text prompts. CLIPSIM projects text and images to the same shared space and evaluates the similarity of their embeddings.

\vspace{5pt}
\textbf{Inference Time:} We test inference time per image on one NVIDIA A100 80G to evaluate efficiency. Results are averaged over $20$ random runs with batch size $1$.

\subsection{Comparison with SD-based Methods}
\label{sec:compare_sd}

We compare \ModelName with the very recent SOTA model ControlNet~\cite{controlnet23} and T2I-Adapter~\cite{t2i23}. 
To ensure fairness, we report results trained on a subset of our proposed \DatasetNameLAION, including 0.2 million (0.2M) images with a data distribution similar to ControlNet and T2I-Adapter's training datasets in Table \ref{tab:all_metrics}.

The superiority of \ModelName~in pose controllability is validated by its remarkable performance in pose-related metrics. Compared with the best results among ControlNet and T2I-Adaptor, \ModelName(0.2M) shows a 34.8\% to 109.1\% performance boost on pose accuracy (e.g., AP, AP(m), and PCE). 
A combination of better condition injection and the \LossName~leads to such performance enhancements. The results interpret the inevitable conflicts in ControlNet stated in Section~\ref{sec:double_sd_conflict}. As indicated previously, with such conflicts, ControlNet may frequently be disrupted by the negative features that exist in the frozen branch, and fail to faithfully render the given pose. Instead, the native generation process and the \LossName of \ModelName simplify the pose guidance and ensure generation quality. Figure \ref{fig:teaser} \uppercase\expandafter{\romannumeral1} and Figure \ref{fig:teaser} \uppercase\expandafter{\romannumeral3} further demonstrate \ModelName's expertise in handling challenging poses. 

Specifically, since text prompts barely indicate the position (corresponding to the metric AP) and the size (corresponding to the metric AP(m)) of each person-to-generate, there undoubtedly exists a large behavior gap between the text-guided frozen branch and the pose-aware trainable branch in ControlNet and T2I-Adapter. This is validated by the almost zero scores of SD on AP and AP(m). Accordingly, ControlNet and T2I-Adapter undoubtedly face severe conflicts related to such properties. Unfortunately, the conflicts aggravate with the increasing complexity of the conditions or the scenarios (e.g., see Figure \ref{fig:teaser}  \uppercase\expandafter{\romannumeral4}). This is validated by their even worse performance on multi-person generation than SD reflected by PCE. 

Moreover, \ModelName infers much faster compared to ControlNet and T2I-Adapter thanks to its single branch design. Although T2I-Adapter uses a more efficient trainable branch than ControlNet, introducing an additional condition learning branch is still time-consuming. The compression of its condition learning branch also leads to quality decline compared with ControlNet, as shown in Table~\ref{tab:all_metrics}.

The disruption in ControlNet may be trivial for CAP, as the pose information given by the text prompts and the skeletal images are likely to be similar (e.g., the pose indicated by the text 'standing' may be very similar to the actual skeleton of a person standing). Therefore, SD can reach a certain level of CAP even without pose conditions, and the improvement of \ModelName~on CAP is relatively small. 

For image quality, the three models show similar FID and KID, indicating that they all manage to preserve SD's basic abilities of image generation and text comprehension. \ModelName~achieves such performance by concentrating on specific human regions in fine-tuning, while ControlNet and T2I-Adapter achieve this via their frozen SD branches. SD has relatively the worst style-wise FID and KID scores, indicating the disqualification of text information in guiding high-quality and diverse human-centric image generation.

However, both \ModelName and ControlNet show a performance decline in text-image consistency. Such negligible yet existent degradation lies in the potential inconsistency between the text and the pose conditions. This implies that both \ModelName~and ControlNet assign higher priority to the pose over the text in the generation.

\subsection{Comparison with GAN-based Methods}
\label{sec:compare_gan}

This section shows the incapability of previous GAN-based HIG methods on precise and diverse pose control. We compare Neural Texture Extraction and Distribution (NTED~\cite{nted22_gan}) and Text-Induced Pose Synthesis (TIPS~\cite{tips22}), which are both textless pose-guided real-scene image generation methods. For NTED and TIPS, we use images randomly selected from DeepFashion~\cite{deepfashion} as source image inputs and the skeleton maps from the validation set of \DatasetNameHumanArt as pose conditions. For \ModelName, we use text and images from the validation set of \DatasetNameHumanArt.

\begin{figure}[htbp]
    \centering
    \includegraphics[width=0.85\linewidth]{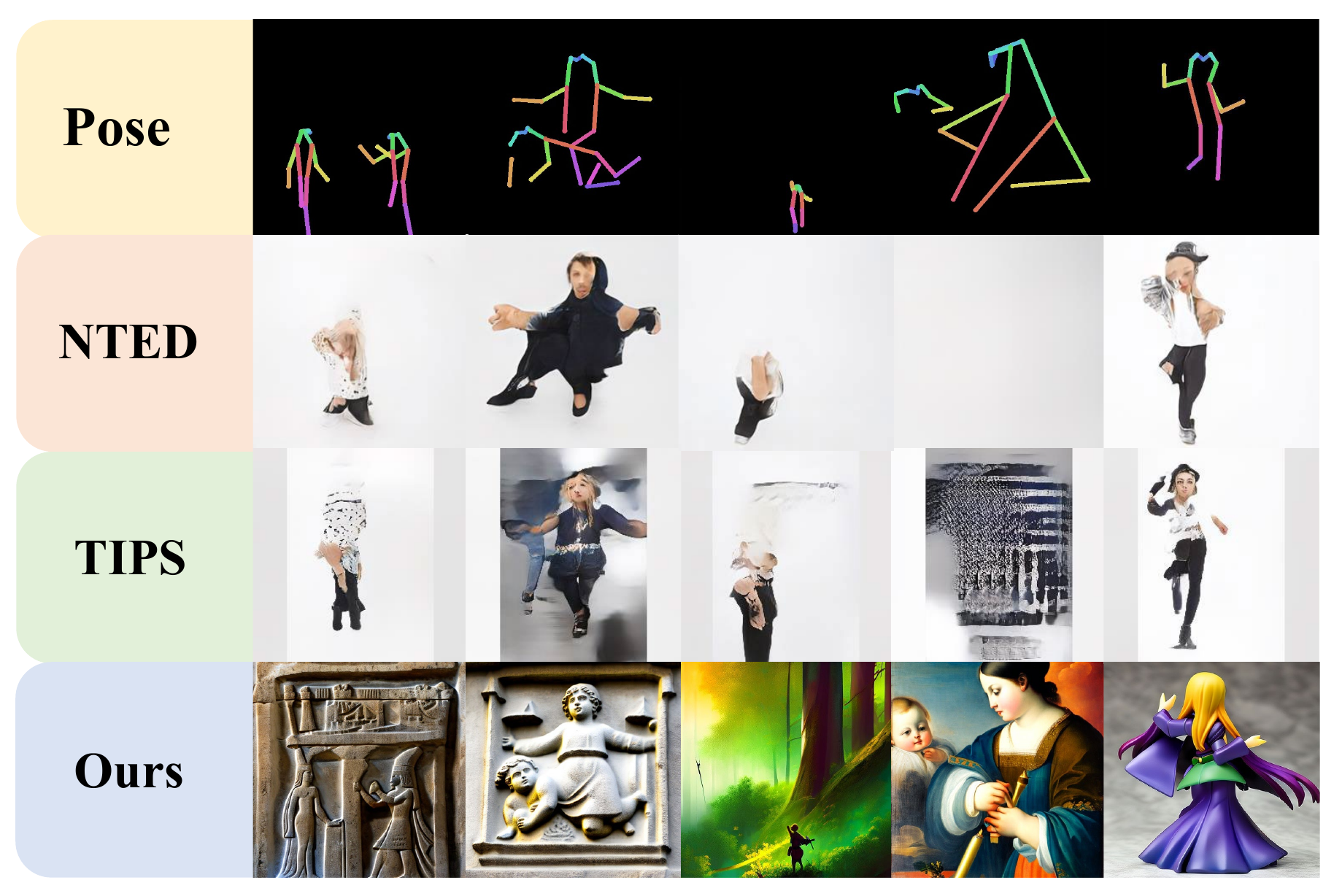}
    \caption{\textbf{Examples generated by previous methods} NTED and TIPS given unconventional pose conditions. 
    }
\label{fig:eg_nted_tips}
\vspace{-0.4cm}
\end{figure}

As shown in Figure~\ref{fig:eg_nted_tips}, NTED and TIPS easily fail given unconventional pose conditions. Specifically, NTED and TIPS have an AP score of $2.79$ and $17.65$ with untrained poses as input. Thus, we can conclude that previous GAN-based HIG methods are not qualified for open-scenario poses. This reflects the significance of \ModelName~with precise pose control and multi-scenario generation ability.

\vspace{-0.25cm}

\subsection{Ablation Study}
\label{sec:ablation}
\vspace{-0.1cm}
In this section, we demonstrate that apart from the better condition injection that allows for a more native generation and avoids conflicts, the proposed \LossName~also contributes to the better performance of \ModelName. Also, we explore how the proposed training datasets and the number of training iterations influence the final results. Unless otherwise stated, the model is trained on Union for around 300 GPU hours (9.5w iterations).

\textbf{Impact of the heatmap-guided denoising loss.} As shown in Table~\ref{tab:ablation_loss}, adding the \LossName helps the back-propagation focus on optimizing weights more related to human generation. This further leads to more precise human pose guidance and thus boosts the AP score from $30.63$ to $32.66$. Meanwhile, focusing on the human generation improves the background's preservation. It thus safeguards the non-human-associated image information to be more related to text descriptions, which increases CLIPSIM from $32.55$ to $32.98$.

\begin{table}[htbp]
\centering
\begin{tabular}{l|cc|c}
 \toprule
   \textbf{Model}           & \textbf{AP}  $\uparrow$   & \textbf{PCE} $\downarrow$ & \textbf{CLIPSIM} $\uparrow$ \\ \midrule
 w/ proposed loss  & \textbf{32.66} & \textbf{1.56}     & \textbf{32.99}   \\
 w/o proposed loss & 30.63 & 1.57     & 32.55   \\  \bottomrule
\end{tabular}
\caption{\textbf{Ablation on the loss function.}}
\label{tab:ablation_loss}
\vspace{-0.4cm}
\end{table}

\begin{figure}[htbp]
    \centering
    \includegraphics[width=1\linewidth]{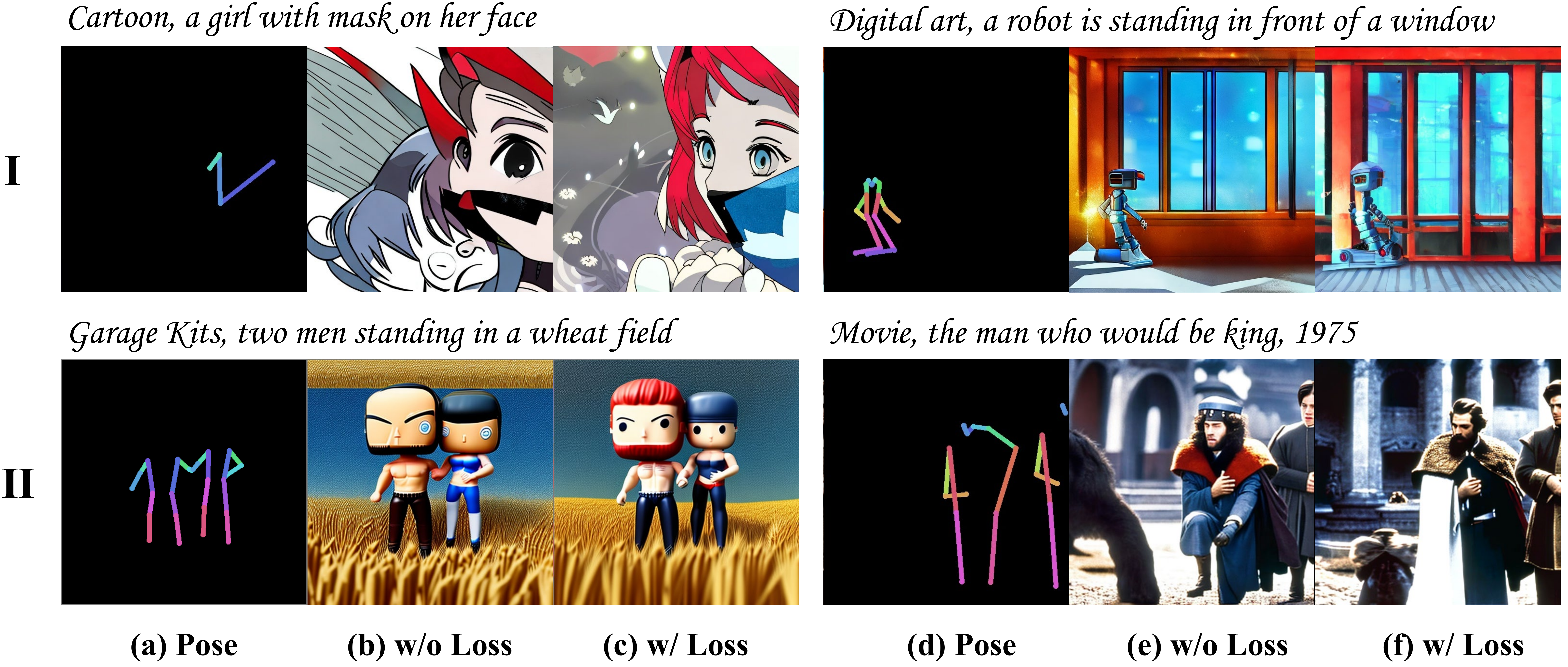}
    \caption{\textbf{Visualization of generated results w/ and w/o the \LossName.} 
    } 
    
\label{fig:add_loss}
\vspace{-0.3cm}
\end{figure}

Figure \ref{fig:add_loss} shows qualitative visualizations of the impact of the \LossName. The \LossName contributes to more precise pose controllability (\uppercase\expandafter{\romannumeral2}(c), \uppercase\expandafter{\romannumeral2}(f)), better human detail fidelity (\uppercase\expandafter{\romannumeral1}(c), \uppercase\expandafter{\romannumeral2}(c)), improved text-image consistency (\uppercase\expandafter{\romannumeral1}(c), \uppercase\expandafter{\romannumeral2}(f)), and enhanced background quality (\uppercase\expandafter{\romannumeral1}(c), \uppercase\expandafter{\romannumeral1}(f), \uppercase\expandafter{\romannumeral2}(c), \uppercase\expandafter{\romannumeral2}(f)). Notably, \ModelName also generates remarkable results on humanoid figures like robots and animals (\uppercase\expandafter{\romannumeral1}(f)).

\vspace{-0.1cm}

\textbf{Impact of training datasets.}  To show the validity of the proposed datasets, we provide results on three training dataset settings. As shown in Table~\ref{tab:ablation_dataset}, using \DatasetNameGHI alone can guarantee the generation with the most accurate human numbers and stronger text-image consistency. This is primarily owing to \DatasetNameGHI's better data distribution alignment with SD. However, due to the absence of real images, results generated by the model trained with \DatasetNameGHI show low generation quality (e.g., blurred human limbs, unrealistic human structure), resulting in a low AP score. Training on \DatasetNameLAION can achieve relatively more satisfactory AP results. Compared with \DatasetNameLAION, combining all datasets to train a model can further improve AP performance, and obtain better trade-offs among PCE and CLIPSIM due to the increase in diversity.

\begin{table}[htbp]
\centering
\begin{tabular}{l|cc|c}
 \toprule
   \textbf{Dataset}           & \textbf{AP}  $\uparrow$   & \textbf{PCE} $\downarrow$ & \textbf{CLIPSIM} $\uparrow$ \\ \midrule
Union  & \textbf{32.66} & 1.56     & 32.99   \\
\DatasetNameGHI & 23.66 & \textbf{1.50}     & \textbf{33.61}   \\
\DatasetNameLAION  & 31.93 & 1.60     & 32.98   \\  \bottomrule
\end{tabular}
\caption{\textbf{Ablation on training datasets.} Details of the datasets are introduced in Section~\ref{sec:dataset_process}}
\label{tab:ablation_dataset}
\vspace{-0.6cm}
\end{table}

\textbf{Impact of training iterations.} Fine-tuning iterations of \ModelName have a significant impact on generation results. In our experiments, the performance increases and then fluctuates at around 95K iterations (around 300 GPU hours). Table~\ref{tab:ablation_step} shows the training results for 50K, 95K, and 150K iterations.

\begin{table}[htbp]
\centering
\begin{tabular}{l|cc|c}
 \toprule
   \textbf{Iterations}           & \textbf{AP}  $\uparrow$   & \textbf{PCE} $\downarrow$ & \textbf{CLIPSIM} $\uparrow$ \\ \midrule
50K & 25.47 & 1.58     & 32.39   \\
95K   & \textbf{32.66} & 1.56     & \textbf{32.99}   \\
150K  & 32.17 & \textbf{1.55}     & 32.51   \\  \bottomrule
\end{tabular}
\caption{\textbf{Ablation on fine-tuning iterations.}}
\label{tab:ablation_step}
\vspace{-0.6cm}

\end{table}

\section{Conclusion}

In this work, we have presented a new framework, named \ModelName, based on pre-trained SD for highly precise pose and text-conditioned human image generation. To concentrate on the generation of foreground humans and preserve pre-trained SD's generation ability, we proposed a novel \LossName. Moreover, we introduced large-scale human-centric datasets containing over 2M text-image-pose pairs for multi-scenario human-centric generative learning. Finally, we compared \ModelName with previous models on a series of evaluation metrics covering image quality, pose accuracy, and text-image consistency. Results demonstrate the effectiveness and efficiency of our proposed method and datasets.

Although the proposed \ModelName improves controllability for HIG, it still encounters certain limitations. (1) \ModelName still easily fails in image generation with extremely crowded scenes and complex/rare actions. (2) Despite being filtered, the large-scale text-image-pose pairs we trained on may still contain social biases and violent content. (3) The evaluation system is not yet comprehensive and complete. 

In our future work, we plan to address the above limitations for better controllable HIG. 
Finally, we hope that this work can motivate future research with a focus on HIG for higher controllability, richer human scenarios, more conditions, and better image quality.


\section*{Acknowledgements}

\vspace{-0.1cm}

This work was supported in part by the Shenzhen-Hong Kong-Macau Science and Technology Program (Category C) of the Shenzhen Science Technology and Innovation Commission under Grant No. SGDX2020110309500101 and in part by Research Matching Grant CSE-7-2022.

\clearpage

\appendix

\section*{Supplementary Materials}

This supplementary material presents more details and additional results not included in the main paper due to page limitation\footnote{Code and datasets will be publicly available for further research}. The list of items included are:

\begin{itemize}

\vspace{-0.2cm}
    \item Experimental details in Sec.~\ref{sec: experimental details}.

\vspace{-0.2cm}
    \item More Quantitative results in Sec.~\ref{sec: quantitative results}.
\vspace{-0.2cm}
    \item More Qualitative results in Sec.~\ref{sec: qualitative results}.
\vspace{-0.2cm}
    \item Future Work in Sec.~\ref{sec: future work}.
\end{itemize}

\section{Experimental Details}
\label{sec: experimental details}

We first provide details of the evaluation metrics in Sec.~\ref{sec: implementation details}, the training details of \ModelName in Sec.~\ref{sec:supp_a2}, and detailed explanations of the \LossName in Sec.~\ref{sec: implementation heatmap}. 

\subsection{Evaluation Metrics}
\label{sec: implementation details}

Details of the evaluation metrics are as follows:
\begin{itemize}
    \item \textbf{Image Quality:}
    
    The calculation of Fr\'echet Inception Distance (FID~\cite{fid}) is given by:
    
\vspace{-0.5cm}

    \begin{equation}
    \small
    FID=\left| \mu -\mu _w \right|+tr\left( \Sigma +\Sigma_w-2\left( \Sigma\Sigma_w \right) ^{\frac{1}{2}} \right) 
    \end{equation}

\vspace{-0.3cm}

    where $\mathcal{N}(\mu, \Sigma)$ is the multivariate normal distribution estimated from Inception v3~\cite{incep} features calculated on \DatasetNameHumanArt and $\mathcal{N}(\mu_w, \Sigma_w)$ is the multivariate normal distribution estimated from Inception v3 features calculated on generated (fake) images. We use Inception v3 with a feature layer of $64$ by default.

    The calculation of Kernel Inception Distance (KID~\cite{kid}) is given by:

\vspace{-0.5cm}

    \begin{equation}
        KID = MMD(f_{real}, f_{fake})^2
    \end{equation}

    where $MMD$ is the maximum mean discrepancy and $I_{real}$, $I_{fake}$ are extracted features from \DatasetNameHumanArt and generated images. 

    The KID and FID are calculated on each scenario in \DatasetNameHumanArt's validation set. Then, we average the results of all scenarios to get the final results. Although the validation set of \DatasetNameHumanArt is relatively small (3,750 images) for FID/KID calculation, given constraints on pose and scenario, we believe they can reflect the quality of generated images to some extent.

    \item \textbf{Pose Accuracy:}

    \begin{figure}[htbp]
    \vspace{-0.2cm}
        \centering
        \includegraphics[width=0.6\linewidth]{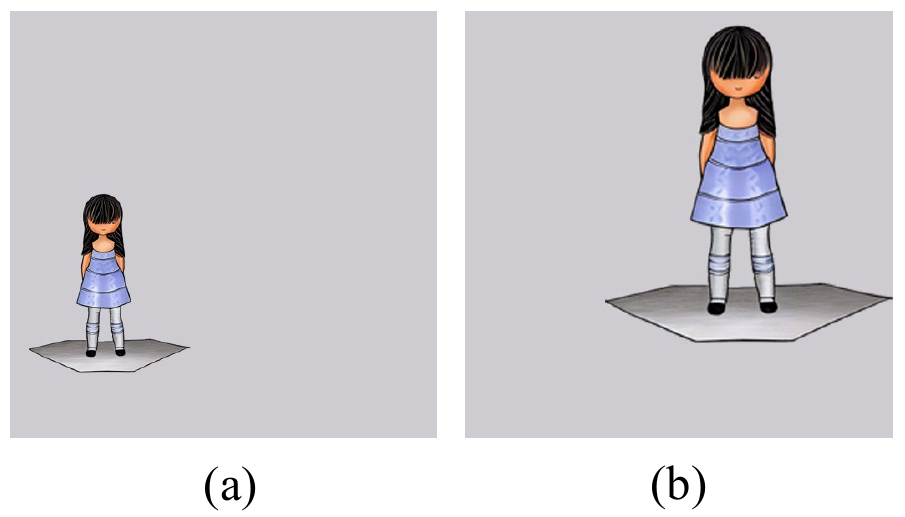}
        \vspace{-0.2cm}
        \caption{Example of calculating the AP and CAP between (a) and (b) images and obtaining an AP of $0$ but a CAP of $1$.
        } 
        \vspace{-0.5cm}
    \label{fig:ap_cap}
    \end{figure}

    A thorough explanation of the pose accuracy metrics is provided in the main paper. We further explain the difference between distance-based average precision (AP) and pose cosine similarity-based AP (CAP). As shown in Figure \ref{fig:ap_cap}, humans in (a) and (b) have the same pose but non-overlapping positions. Take the pose of (a) as the target pose. The AP of (b) is $0$, but the CAP is $1$. Thus, CAP can eliminate the influence of position and focus on the similarity between the two poses. A combination of CAP and AP can better demonstrate the pose controllability over the generated person's pose and position.

    \item \textbf{Text-image Consistency:}

    We employ CLIP-ViT-base-path16 to extract text and image features, using a ViT-B/16 Transformer as the image encoder and a masked self-attention Transformer as the text encoder.
    
    \end{itemize}

\subsection{Prompt Engineering of the GHI Dataset}

This work emphasizes the essence of the multi-scenario human-centric image generation task with precise pose control. The quality and scale of the training datasets are crucial. To better control the quality and content of image generation, we design a set of systematic rules for generating prompts.
Prompt engineering ~\cite{oppenlaender2022prompt,liu2022design} is one method for increasing the quality of text-to-image models. 

\begin{figure}[htbp]
        \centering
        \vspace{-0.1cm}
        \includegraphics[width=0.72\linewidth]{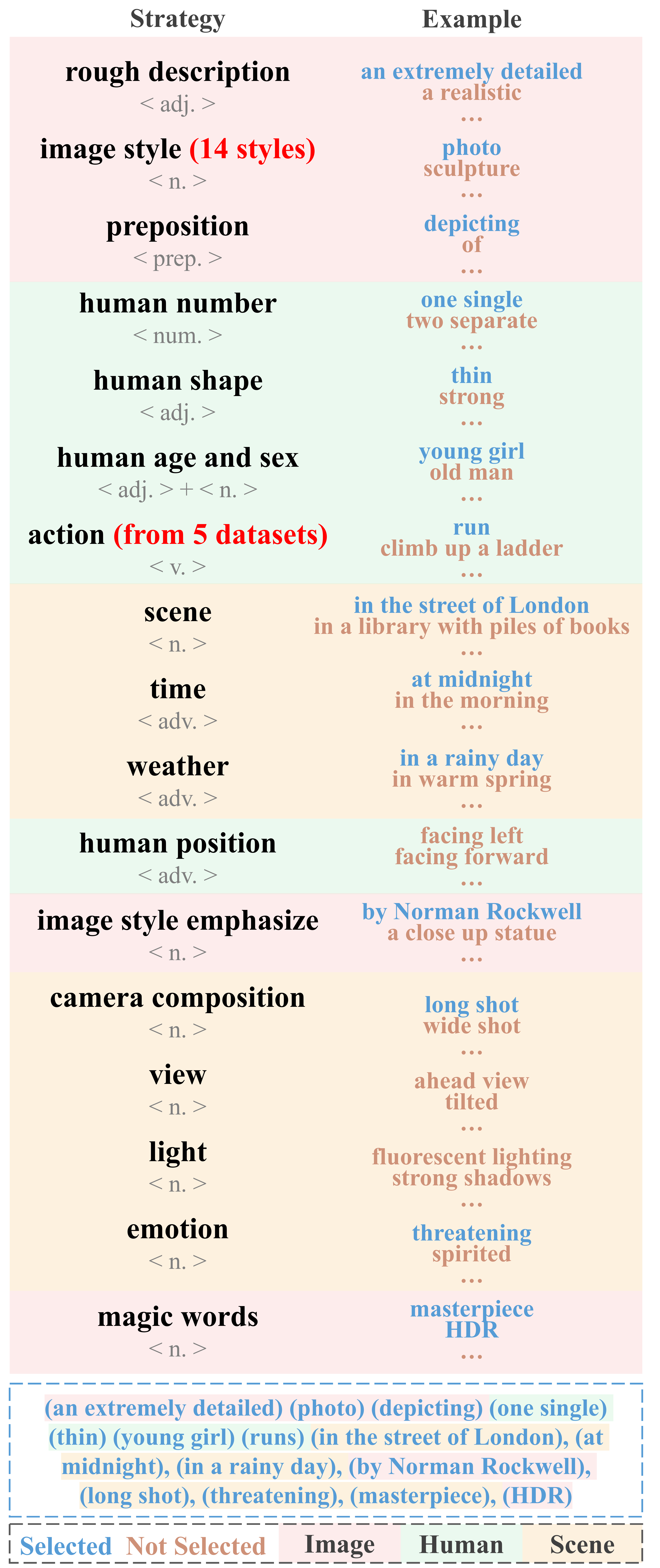}
        \caption{Components of prompts in GHI. Note that the order of prompts has an impact on the generation results.
        } 
        \vspace{-0.4cm}
    \label{fig:prompt}
    \end{figure}

In generating the GHI dataset with high-quality, diverse, and human-centric text and the corresponding images, we take advantage of prompt engineering to design large quantities of unrepeatable prompts with a high guarantee of image quality. Specifically, the prompt is composed of 18 parts to describe three main components, like image, human, and scene. Figure \ref{fig:prompt} includes a comprehensive description of the whole image (e.g., the image style), human features (e.g., the human number, shape, and action), and the background scene (e.g., time, weather, and camera settings). Different parts have distinct selection probabilities and numbers, culminating in a variety of rich and diverse prompts. To ensure the diversity of image styles, we adopt $14$ different styles referring Human-Art\cite{humanart} to cover as many image styles as possible, including photo, garage kits, relief, statue, kids drawing, mural, oil painting, sketch, stained glass, ukiyoe, cartoon, digital art, ink painting, and watercolor. To ensure the diversity of human action, we collect $6826$ different human actions referring recent popular human action datasets BABEL~\cite{BABEL}, NTU RGB+D 120~\cite{liu2019ntu}, HuMMan~\cite{cai2022humman}, HAA500~\cite{chung2021haa500}, and HAKE-HICO~\cite{li2019hake}.

Based on our designed prompts, the used stable diffusion model can generate images of great diversity and quality. However, they may still fail to faithfully respect human structures and generate missing, redundant, replaced body parts or wrong human numbers, which are key to the human-centric image generation task. KPE~\cite{kpe22_vae} clarifies the validity of using pose estimators to judge the correctness of both generated body structure and human number. When a pose estimator is fed images with an unreasonable body structure, it typically assigns the incorrect body component to an additional human, resulting in an inconsistent human count. Accordingly, we use the pre-trained pose estimator HigherHRNet to determine whether the estimated human number equals the given number (1-3 humans with a proportion of 7:2:1). We finally reserve $4$ images with a correct human number for each prompt in GHI.

\subsection{The Training Details of HumanSD}
\label{sec:supp_a2}

We train \ModelName with $4$ NVIDIA A100 Tensor Core GPUs. By default, we train each model with a batch size of $4$ for about $3$ days, around $95,000$ iterations and $300$ GPU hours. Different from ControlNet, which receives a sudden convergence around the long-lasting $6100_{th}$ iteration, \ModelName shows a fast but smooth convergence from the $0_{th}$ iteration to the $600_{th}$ iteration. Although \ModelName can generate humans with corresponding pose conditions after only around $600$ iterations, training with more iterations leads to better performance. However, training with more than $95,000$ iterations does not improve metrics in our experiments. The results are also shown in Table 4.

\subsection{Detailed Implementation of the Heatmap-Guided Denoising Loss}
\label{sec: implementation heatmap}

As explained in the main paper, the vanilla LDM\cite{diffusion2015} has a loss function:

\vspace{-0.5cm}

\begin{equation}
\small
L_\text{LDM}=\mathop \mathbb{E} \limits_{t,z,\epsilon}\left[ \left\| \epsilon -\epsilon _{\theta}\left( \sqrt{\bar{\alpha}_t}z_0+\sqrt{1-\bar{\alpha}_t}\epsilon, c, t \right) \right\| ^2 \right]
\label{eq:ldm_loss}
\end{equation}

\vspace{-0.2cm}

To obtain a difference map to be recognized by the pose estimator, we feed $\epsilon -\epsilon _{\theta}$ into the VAE decoder of Stable Diffusion (SD) and get:

\vspace{-0.2cm}

\begin{equation}
M=VAE_{decoder}\left( \epsilon -\epsilon _{\theta} \right) 
\end{equation}

\vspace{-0.1cm}

where $M$ is the difference map of noise difference.

We use a bottom-up pose estimator HigherHRNet~\cite{cheng2020higherhrnet} pre-trained on MSCOCO\cite{coco} and \DatasetNameHumanArt for heatmap estimation of $Difference\,\,Map$. The bottom-up pose estimator shows better performance on blurred difference maps. Moreover, the estimation can become more inclusive by combining MSCOCO and \DatasetNameHumanArt in training. We determine the heatmap by:

\vspace{-0.2cm}

\begin{equation}
H=F\left( M \right) 
\end{equation}

\vspace{-0.1cm}

where $H \in\mathbb{R}^{\textbf{h}\times \textbf{w}\times \textbf{k}}$ is the heatmap matrix with height $\textbf{h}$ and width $\textbf{w}$. $\textbf{k}$ is the human joint number. $F$ is the heatmap estimator. For ease of calculation, we then sum $H$ across the joint dimension to generate a single heatmap.

The larger the difference between the output noise of the UNet $\epsilon_\theta$ and ground-truth noise $\epsilon$ is, the more noticeable the human figure will be in $M$, and the larger value $H$ will have at the corresponding joint positions. Therefore, to get the heatmap mask, we set $0.1$ as the empirical threshold on $H$ to form heatmap mask $H_M$. Then, we pass $H_M$ back to the VAE encoder to get the heatmap embedding.

\vspace{-0.2cm}

\begin{equation}
H_E=VAE_{encoder}\left( H_M \right) 
\end{equation}

\vspace{-0.1cm}

where $H_E$ is the heatmap embedding.

Finally, the weighted loss is calculated as follows:

\vspace{-0.2cm}
\begin{equation}
\small
L_\text{h}= \mathop \mathbb{E} \limits_{t, z, \epsilon}\left[   \left\| W_a \cdot  \left( \epsilon -\epsilon _{\theta}\left( \sqrt{\bar{\alpha}_t}z_0+\sqrt{1-\bar{\alpha}_t}\epsilon, c, t \right)\right) \right\| ^2 \right] 
\end{equation}

\vspace{-0.1cm}

where $ W_a=w\cdot H_E+1$. $w$ is set $0.05$ by default.

\section{Quantitative results}
\label{sec: quantitative results}

This section reports more quantitative results to comprehensively assess pose controllability in different image scenarios and when inputting pose conditions with different human numbers.

\begin{figure}[htbp]
    \centering
    \vspace{-0.2cm}
    \includegraphics[width=1\linewidth]{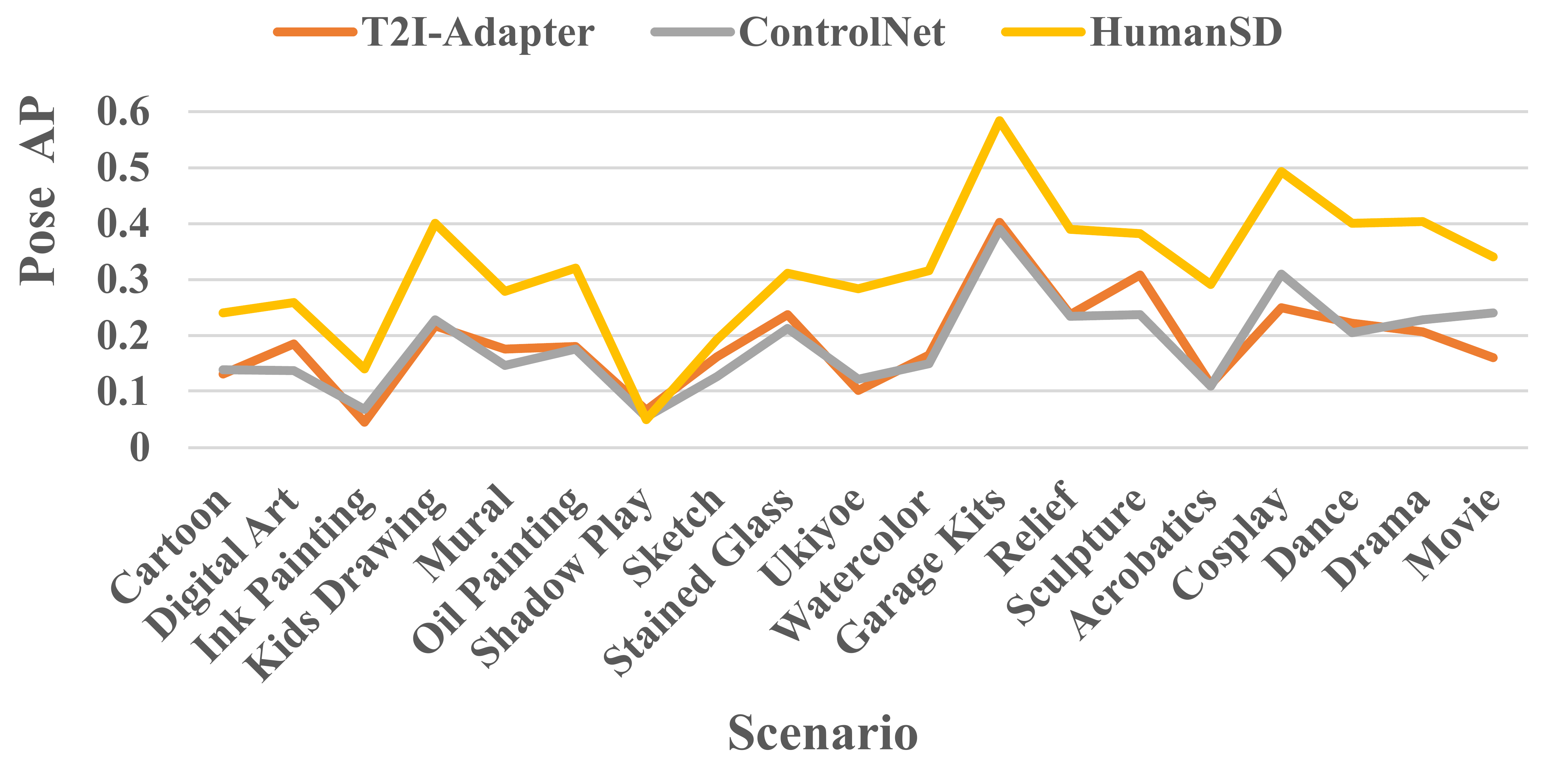}
    \vspace{-0.5cm}
    \caption{Comparisons of the pose average precision (Y-axis) on different scenarios (X-axis).
    } 
    \vspace{-0.3cm}

\label{fig:pose_ap_with_scenario}
\end{figure}

\textbf{Comparisons on different image scenarios.} As shown in Figure \ref{fig:pose_ap_with_scenario}, all methods exhibit comparable patterns in Pose AP throughout different scenarios, with garage kits having the highest AP and shadow play having the lowest AP. This is partially due to the uneven distribution of the training and fine-tuning datasets, where garage kits have a large amount of data with multi-view and shadow play has only a small number of images. Natural scenarios such as cosplay also have a comparatively higher AP, which again reflects the differential distribution of the dataset. However, due to the complexity of pose conditions, natural scenes such as acrobatics and dance do not show a high AP. 

\ModelName consistently displays higher AP in all but the shadow play scenario. The lack of shadow play images in the dataset used to train T2I-Adapter, and ControlNet is most likely to blame for this. As unaware of the shadow play scenario, these two methods are more likely to produce images with real people (as shown in the first figure of Figure 1 \uppercase\expandafter{\romannumeral2} in the main paper) as  a replacement for shadow play. By cheating the pose accuracy criteria, a falsely high AP score is obtained. From the comparison of different image scenarios, we can lead to two conclusions: (1) The wide variation of APs across different scenarios indicates that the potential challenges in various scenarios are different, and that evaluating and generating multiple scenario images is still challenging. (2) \ModelName outperforms ControlNet and T2I-Adapter in all scenarios, especially cosplay, with a boost of 18.2\% to ControlNet and 24.1\% to T2I-Adapter.

\begin{figure}[htbp]
    \centering
        \vspace{-0.2cm}
    \includegraphics[width=1\linewidth]{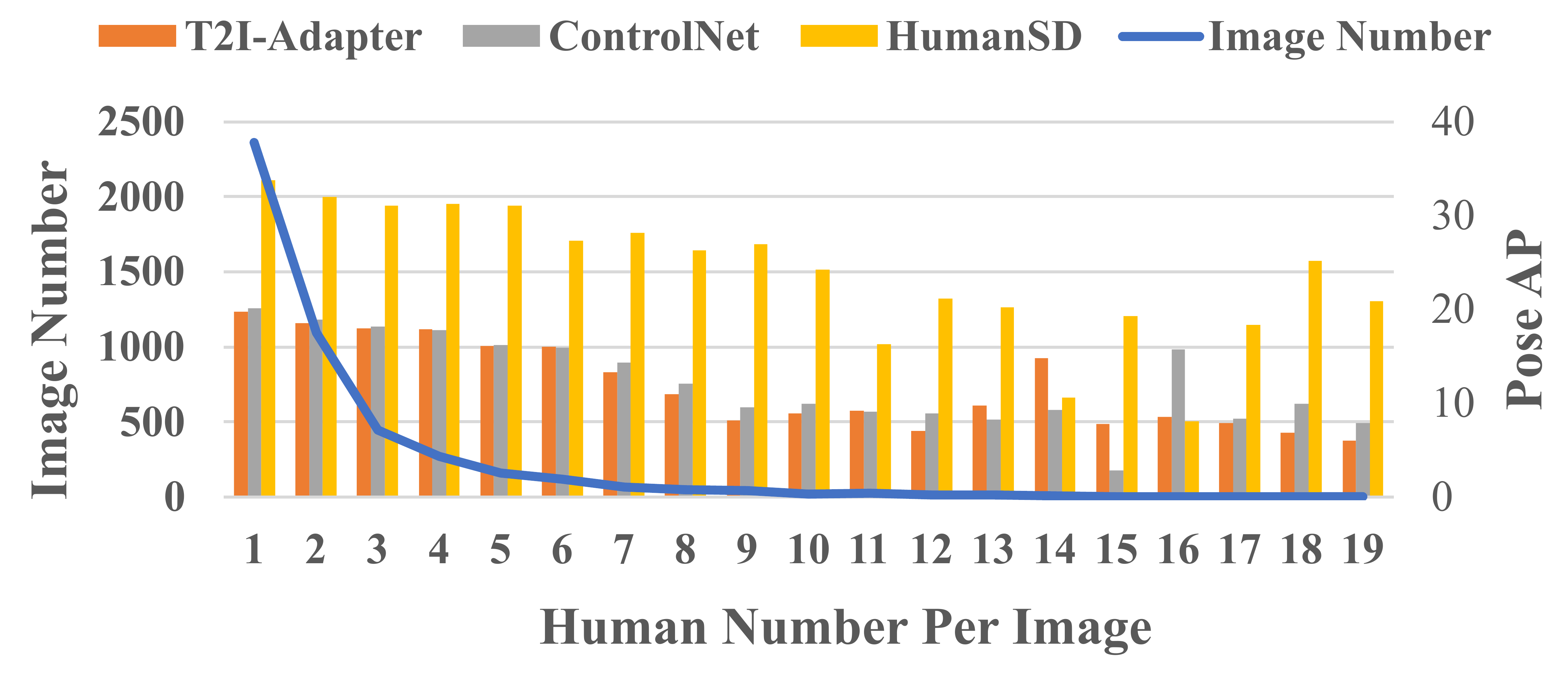}
    \vspace{-0.5cm}
    \caption{Comparisons of pose average precision (AP) (right Y-axis with the histogram) with 1 to 19 human number per image (X-axis) in the validation set of \DatasetNameHumanArt. The image number statistic is shown in the left Y-axis with the blue curve to state the total sample size.
    } 
        \vspace{-0.3cm}
\label{fig:pose_ap_with_number}
\end{figure}

\textbf{Comparisons on human numbers per image.} As shown in Figure~\ref{fig:pose_ap_with_number}, the Pose AP value tends to decline with the increase of human number in a single image, which shows the difficulty in generating images with multiple persons. \ModelName can still retain a high pose AP as the number of humans increases, demonstrating its ability to generate the multi-human image. Moreover, \ModelName shows a unified better AP score among images with $1$-$13$ humans than ControlNet and T2I-Adapter. The limited number of images with $14$-$19$ humans lead to fluctuation of Pose AP, but \ModelName still shows a relatively better result.

\vspace{-0.1cm}

\section{Qualitative Results}
\label{sec: qualitative results}
\vspace{-0.1cm}

\begin{figure*}[htbp]
    \centering
    \vspace{-0.5cm}
    \includegraphics[width=0.95\linewidth]{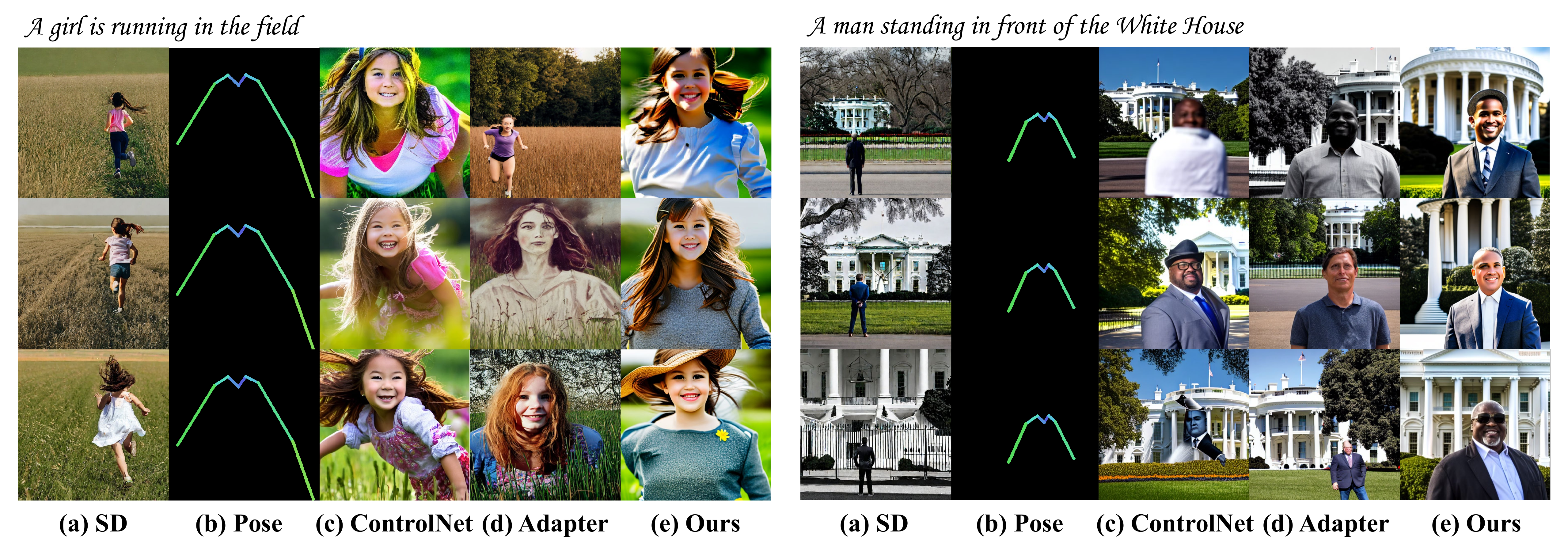}
    \caption{\textbf{Natural Human Scene - Half Body}. (a) a generation by the pre-trained text-guided stable diffusion (SD)~\cite{ldm22}, (b) pose skeleton images as the condition to ControlNet, T2I-Adapter and our proposed \ModelName, (c) a generation by ControlNet~\cite{controlnet23}, (d) a generation by T2I-Adapter~\cite{t2i23}, and (e) a generation by \ModelName (ours). ControlNet, T2I-Adapter, and \ModelName receive both text and pose conditions. We use three different seeds (the three rows) to generate diverse images. 
    } 
    \vspace{-0.5cm}
\label{fig:natural_human_scene_normal_poses}
\end{figure*}

\begin{figure*}[htbp]
    \centering
    \vspace{-0.5cm}
    \includegraphics[width=0.95\linewidth]{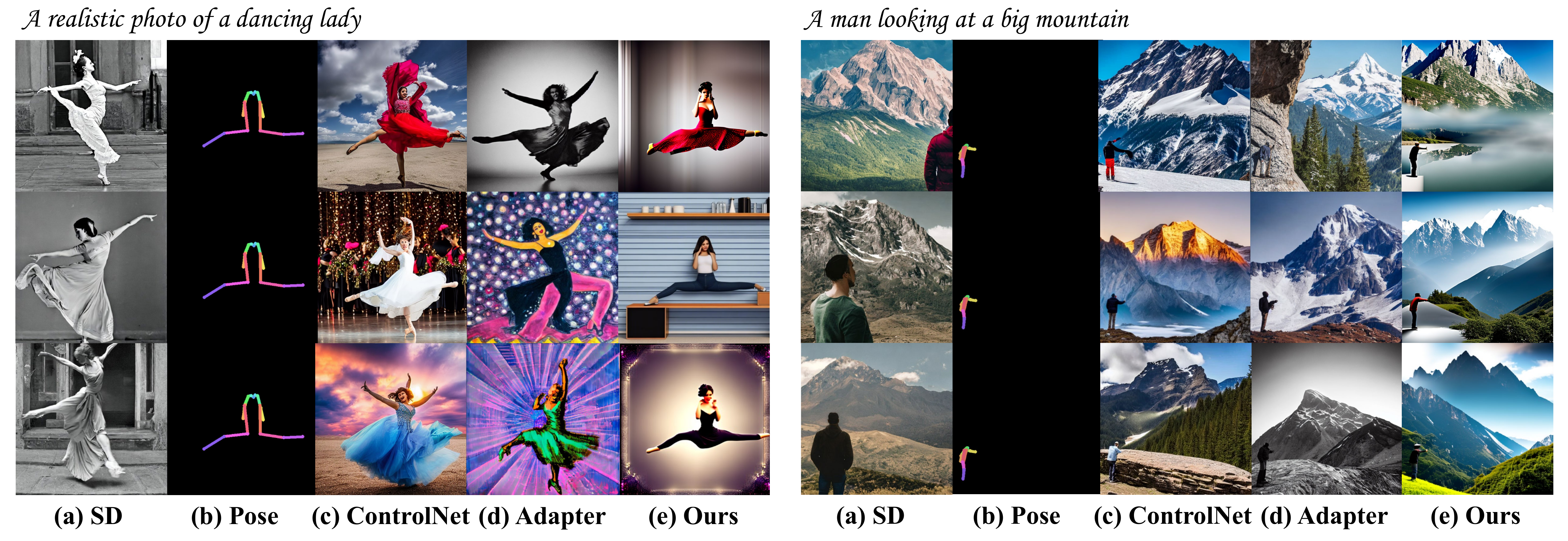}
    \caption{\textbf{Natural Human Scene - Hard Poses/Small Human}. The explanation of (a)-(e) can be found in Figure \ref{fig:natural_human_scene_normal_poses}'s caption.
    } 
    \vspace{-0.5cm}
\label{fig:natural_human_scene_hard_poses}
\end{figure*}

\begin{figure*}[htbp]
    \centering
    \vspace{-0.5cm}
    \includegraphics[width=0.95\linewidth]{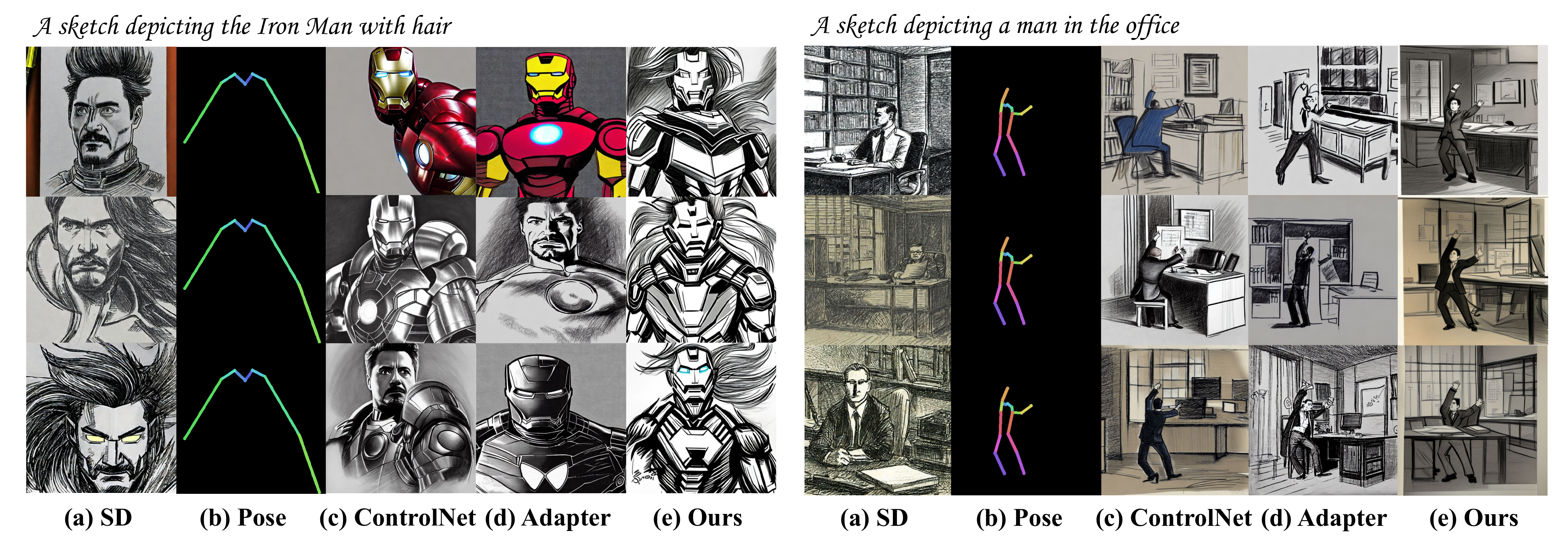}
    \caption{\textbf{Sketch Scene - Text Control/Human Orientation}. The explanation of (a)-(e) can be found in Figure \ref{fig:natural_human_scene_normal_poses}'s caption.
    } 
    \vspace{-0.5cm}
\label{fig:sketch}
\end{figure*}

\begin{figure*}[htbp]
    \centering
    \vspace{-0.5cm}
    \includegraphics[width=0.95\linewidth]{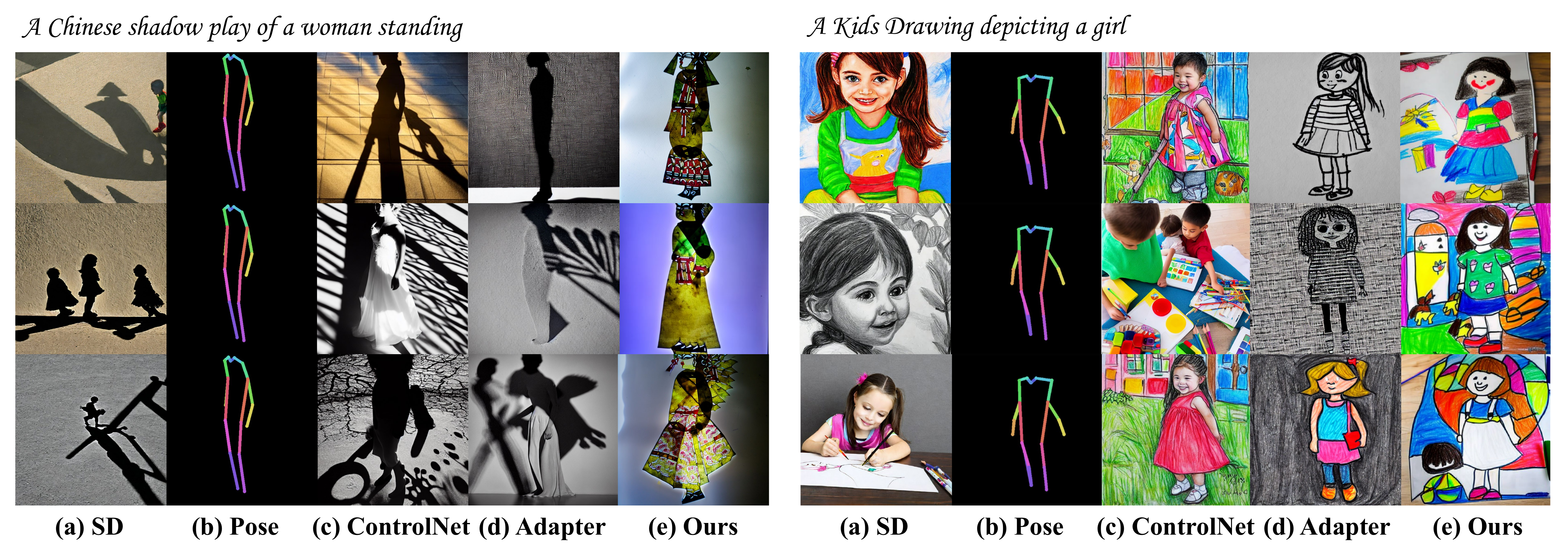}
    \caption{\textbf{Shadow Play / Kids Drawing Scene - Rare Scenes}. The explanation of (a)-(e) can be found in Figure \ref{fig:natural_human_scene_normal_poses}'s caption.
    } 
    \vspace{-0.5cm}
\label{fig:shadow_play}
\end{figure*}

\begin{figure*}[htbp]
    \centering
    \vspace{-0.5cm}
    \includegraphics[width=0.95\linewidth]{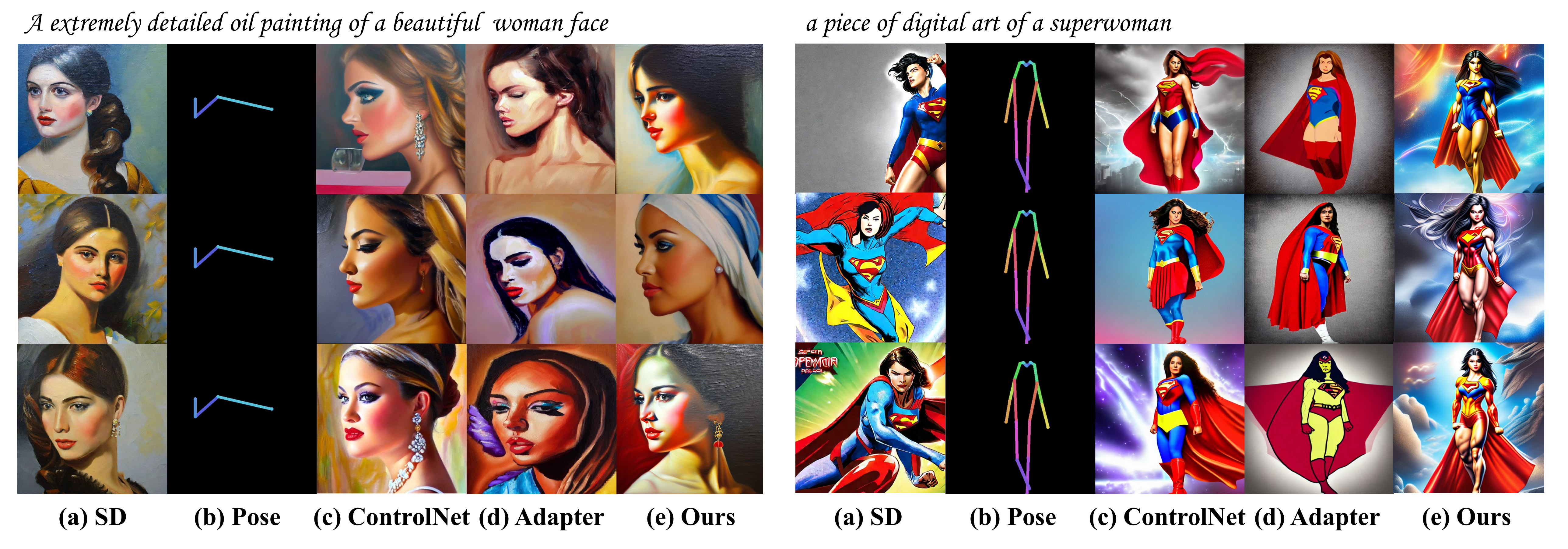}
    \caption{\textbf{Oil Painting / Digital Art Scene - Human Detail}. The explanation of (a)-(e) can be found in Figure \ref{fig:natural_human_scene_normal_poses}'s caption.
    } 
    \vspace{-0.5cm}
\label{fig:oil_painting}
\end{figure*}

More qualitative comparison of ControlNet, T2I-Adapter, and \ModelName is shown from Figure \ref{fig:natural_human_scene_normal_poses} to \ref{fig:oil_painting}, which correspondingly shows the natural human scene with only half body, hard poses / small human in the natural scene, text and human orientation controllability in sketch scene, rare scenes such as shadow play and kids drawing, and human detail in oil painting and digital art. We generate three groups of images with different seeds for each text and pose condition to avoid randomness and show diversity.

\begin{figure*}[htbp]
    \centering
    \vspace{-0.5cm}
    \includegraphics[width=0.95\linewidth]{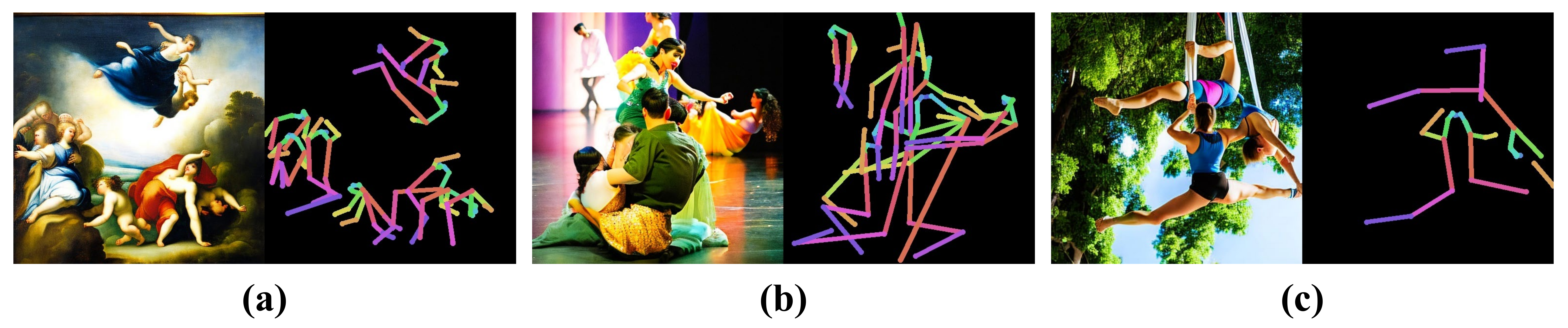}
    \caption{\textbf{Failure cases} on (a)/(b) extremely crowded scenes and (c) complex/rare actions.
    } 

    \vspace{-0.5cm}
    
\label{fig:fail}
\end{figure*}

\vspace{-0.1cm}

\section{Future Work}
\label{sec: future work}
\vspace{-0.1cm}

Although \ModelName has reached a high performance, there are still many issues waiting for exploration. Specifically, future directions include but are not limited to: 
(1) We notice a significant trade-off between whole-body generation and local body part generation. For example, as shown in Figure \ref{fig:oil_painting}, the left image can generate high-fidelity human faces. But when we force the model to generate whole body images in the right image, the facial detail retention shows a huge decline, which is extremely obvious in ControlNet. We leave it to future work for solutions.
(2) \ModelName still fails in extremely crowded scenes and complex/rare actions, as shown in Figure \ref{fig:fail}. Generation models with higher accuracy and faster speed are still in need.
(3) Similar to other generation tasks, the text and pose-guided image generation evaluation system are not yet comprehensive and complete, which entails a lot of randomnesses.
(4) augmentations for complex poses and different orientations of humans. We have noticed that human poses that do not frequently appear (e.g., stand upside down) tend to fail more frequently. Appropriate augmentations may alleviate this problem in the future.

\clearpage
\clearpage
\newpage
\newpage

{
\small
\bibliographystyle{ieee_fullname}
\bibliography{humansd}
}

\end{document}